\documentclass[lettersize,journal]{IEEEtran}
\usepackage{cite}
\usepackage[utf8x]{inputenc} 
\usepackage{amsmath,amssymb,amsfonts}
\usepackage{algorithmic}
\usepackage[table,xcdraw]{xcolor}
\usepackage{float}
\usepackage{graphicx}
\usepackage{longtable}
\usepackage{multirow}
\usepackage{url}
\usepackage{textcomp}
\usepackage{bm}
\usepackage{arydshln}
\begin{document}

\title{A Comprehensive Survey of Convolutions in Deep Learning: Applications, Challenges, and Future Trends}

\author{Abolfazl Younesi\IEEEauthorrefmark{1}, Mohsen Ansari\IEEEauthorrefmark{1}, MohammadAmin Fazli\IEEEauthorrefmark{1}, Alireza Ejlali\IEEEauthorrefmark{1}, Muhammad Shafique\IEEEauthorrefmark{2}, and Jörg Henkel\IEEEauthorrefmark{3}

\thanks{A. Younesi, M. Ansari, A. Ejlali, and M. A. Fazli are with the Department of Computer Engineering, Sharif University of Technology, Tehran, Iran. E-mail: \{abolfazl.yunesi, ansari, ejlali, fazli\}@sharif.edu.}

\thanks{M. Shafique is with the Department of Computer Engineering, New York University (NYU) Abu Dhabi, United Arab Emirates. E-mail: muhammad.shafique@nyu.edu).}

\thanks{J. Henkel is with the Karlsruhe Institute of Technology, Karlsruhe 76131, Germany. E-mail: henkel@kit.edu.}
}


\maketitle

\begin{abstract}
In today’s digital age, Convolutional Neural Networks (CNNs), a subset of Deep Learning (DL), are widely used for various computer vision tasks such as image classification, object detection, and image segmentation. There are numerous types of CNNs designed to meet specific needs and requirements, including 1D, 2D, and 3D CNNs, as well as dilated, grouped, attention, depthwise convolutions, and NAS, among others. Each type of CNN has its unique structure and characteristics, making it suitable for specific tasks. It’s crucial to gain a thorough understanding and perform a comparative analysis of these different CNN types to understand their strengths and weaknesses. Furthermore, studying the performance, limitations, and practical applications of each type of CNN can aid in the development of new and improved architectures in the future. We also dive into the platforms and frameworks that researchers utilize for their research or development from various perspectives. Additionally, we explore the main research fields of CNN like 6D vision, generative models, and meta-learning. This survey paper provides a comprehensive examination and comparison of various CNN architectures, highlighting their architectural differences and emphasizing their respective advantages, disadvantages, applications, challenges, and future trends. 
\end{abstract}

\begin{IEEEkeywords}
Deep learning, DNN, CNN, Machine learning, Vision Transformers, GAN, Attention, Computer Vision, LLM, Large Language Model, Transformer, Dilated Convolution, Depthwise, NAS, NAT, Object Detection, 6D Vision, Vision Language Model 
\end{IEEEkeywords}

\section{Introduction}
\label{sec:introduction}
\IEEEPARstart{I}{N} today's world, as technology continues to evolve, deep learning (DL) has become an integral part of our lives \cite{ref1}. From voice assistants like Siri and Alexa to personalized recommendations on social media platforms, DL algorithms are constantly working behind the scenes to understand our preferences and make our lives easier \cite{ref2}. With advancements in technology, DL is also being used in various fields such as healthcare, finance, and transportation, revolutionizing the way we approach these industries \cite{ref3,ref4,ref5}. As research and development in the field of DL continue to progress, even more innovative applications that will further enhance our daily lives can be expected. DL has ushered in a transformative era in artificial intelligence, empowering machines to assimilate vast datasets and make informed predictions \cite{ref6}\cite{ref8}.  The development of CNNs has received attention among deep learning's significant advancements. Their impact has been felt in some areas, including generative AI, examining medical images, identifying objects \cite{ref9}, and finding anomalies \cite{ref10}. CNNs, constituting a feedforward neural network, integrate convolution operations into their architecture \cite{ref7}\cite{ref11}. These operations enable CNNs to adeptly capture intricate spatial and hierarchical patterns, rendering them exceptionally well-suited for image analysis tasks \cite{ref12}.

However, CNNs are often burdened by their computational complexity during training and deployment, particularly when operating on resource-constrained devices like mobile phones and wearables \cite{ref12}\cite{ref13}.

\begin{table*}[]
\centering
\setlength\extrarowheight{3pt}
\caption{Comparison of existing surveys; +* means conditional cosideration}
\label{tab:existing}
\resizebox{\textwidth}{!}{%
\begin{tabular}{clccccccc}
\hline
\rowcolor[HTML]{C0C0C0} 
{\color[HTML]{000000} \textbf{Ref.}} &
  \multicolumn{1}{c}{\cellcolor[HTML]{C0C0C0}{\color[HTML]{000000} \textbf{Year}}} &
  {\color[HTML]{000000} \textbf{No. of  included studies}} &
  {\color[HTML]{000000} \textbf{Research Questions and Objective}} &
  {\color[HTML]{000000} \textbf{Taxonomy}} &
  {\color[HTML]{000000} \textbf{Datasets}} &
  {\color[HTML]{000000} \textbf{Challanges}} &
  {\color[HTML]{000000} \textbf{Comparison of Simulators}} &
  {\color[HTML]{000000} \textbf{Evaluation}} \\ \hline
\cite{ref117}    & 2023                     & 210 & - & +* & - & -  & + & - \\
\cite{ref118}    & 2021                     & 343 & - & +  & + & +  & + & - \\
\cite{ref119}    & 2022                     & 202 & - & +  & - & +  & - & + \\
\cite{ref120}    & 2020                     & 243 & - & +* & + & +* & - & - \\
Our survey & \multicolumn{1}{c}{2024} & 465 & + & +  & + & +  & + & + \\ \hline
\end{tabular}%
}
\end{table*}
Two principal avenues have emerged to reinforce the energy efficiency of CNNs:
Employing Lightweight CNN Architectures: These architectures are deliberately engineered to achieve computational efficiency without compromising accuracy. For instance, the MobileNet family of CNNs has been meticulously tailored for mobile devices and has demonstrated state-of-the-art accuracy across various image classification Applications \cite{ref13}.

Embracing Compression Techniques: These methods facilitate the reduction of CNN model size and consequently diminish the volume of data transfers between devices. A noteworthy example is the TensorFlow Lite framework, which offers a suite of compression techniques tailored for compressing CNN models for mobile devices \cite{ref14}.

The fusion of lightweight CNN architectures and compression techniques yields a substantial boost in the energy efficiency of CNNs. Training and deploying CNNs on resource-constrained devices become feasible, thereby unlocking novel opportunities for employing CNNs in diverse applications like healthcare, agriculture, and environmental monitoring \cite{ref12}\cite{ref16}.

How different convolutional techniques cater to various AI applications. Convolutions play a fundamental role in contemporary DL architectures and are especially crucial when dealing with data organized in grid-like structures, such as images, audio signals, and sequential data \cite{ref23}. The convolutional operation entails moving a small filter, also known as a kernel, across the input data, performing element-wise multiplications and aggregations. This process extracts essential features from the input data \cite{ref24}. The main significance of convolutions lies in their capability to efficiently capture local patterns and spatial relationships within the data. This localization property makes convolutions highly suitable for tasks like image recognition, as objects can be identified based on their local structures. Additionally, convolutions introduce parameter sharing, which results in a significant reduction in the number of trainable parameters, leading to more efficient and scalable models \cite{ref25}.
\textbf{Existing surveys: }
Previous survey papers on CNN architectures such as \cite{ref118} and \cite{ref120} provided good overviews of popular architectures from a certain period. However, they lacked a clear Research question and objective, evaluation, and challenges based on their design patterns. They mostly discussed architectures chronologically.

Earlier surveys like \cite{ref119} and \cite{ref120} focused on explaining core CNN components and popular architectures up to a certain year. they also lacked research questions and objectives, analysis of datasets, and special types of taxonomy that were not considered complete overviews like large vision models, and large language models, and lacked of multipoint view for challenges.

Previous work discussed the challenges in some specific concepts and applications of CNNs but did not extensively cover the intrinsic taxonomy present in newer CNN architectures. So this caused us to write a survey paper that aims to address the gaps in previous work by proposing a taxonomy to clearly classify CNN architectures based on their intrinsic design patterns rather than release year.

We focus on architectural innovations from 2012 onwards and discuss the recent developments in greater depth than earlier surveys.
Discussing the latest trends and challenges provides an updated perspective for researchers.

\textit{This comprehensive survey of CNN's history, taxonomy, applications, and challenges is needed to accelerate research progress in this domain further.}

In this paper, the key questions we seek to address include:

\begin{itemize}
    \item How do state-of-the-art CNN models like ResNet, Inception, and MobileNet perform on the target hardware compared to constrained baselines? What are the impacts on accuracy, latency, and memory usage?
    \item What techniques like pruning, quantization, distillation, and architecture design can help reduce the model size and computational complexity the most while retaining prediction quality?
    \item How do multi-stage optimization approaches that combine different techniques compare to single methods? Can we achieve better trade-offs between accuracy, latency, and memory?
    \item For a target application like embedded vision, what are the best practices for benchmarking, tuning, and deploying optimized CNN models considering their unique constraints and specifications?
    \item Which pruning and quantization techniques work best for our target application and hardware? How does this compare to baselines?
\end{itemize}

\begin{figure*}[h]
    \centering
    \includegraphics[width=\textwidth]{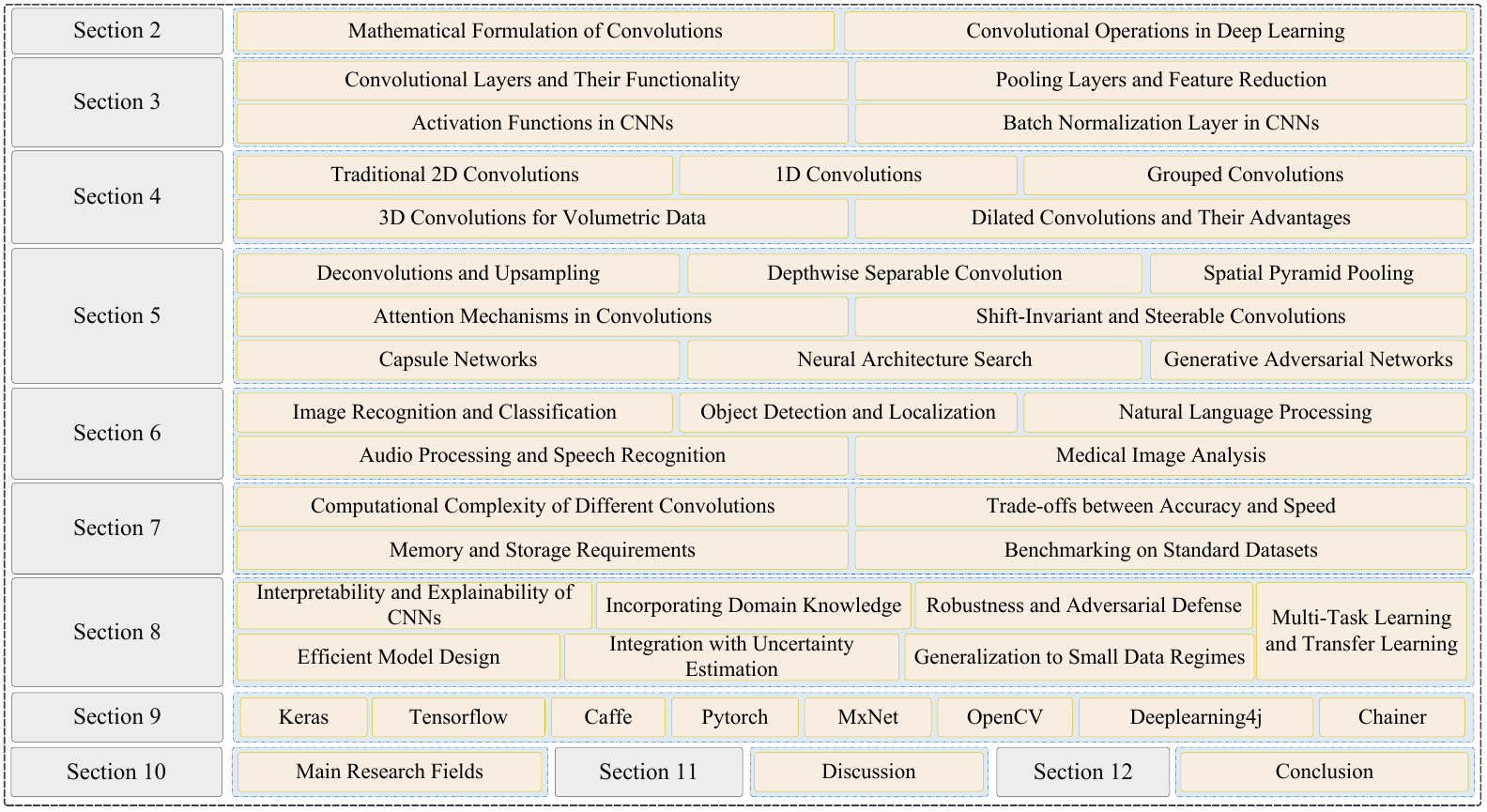}
    \caption{Represents the section-by-section structure of the paper that provides a clear and organized framework for presenting the research findings.}
    \label{fig:structure}
\end{figure*}

Our overview makes several key contributions to the DL and CV communities:
\begin{itemize}
    \item \textbf{Analyzing multiple types of existing CNNs:} The survey provides a comprehensive and detailed analysis of various DL models and algorithms used in CV Applications.
    \item \textbf{Comparing the CNN models with various parameters and architectures:} The overview offers insights into the performance and efficiency trade-offs. 
    \item \textbf{Identifying the strengths and weaknesses of different CNN models:} Aiding researchers in selecting the most suitable model for their specific applications.
    \item \textbf{The overview highlights the challenges and future directions} for further improvement in the fields of DL and computer vision.
    \item \textbf{Exploring the trends in neural network architecture:} This emphasizes the practical application and exciting nature of the advancements.
    \item \textbf{comprehensive overview of the Main research fields:} This covers the primary fields of research that are actively pursued by researchers.
\end{itemize}

\begin{figure}[t]
    \centering
    \includegraphics[width=\columnwidth]{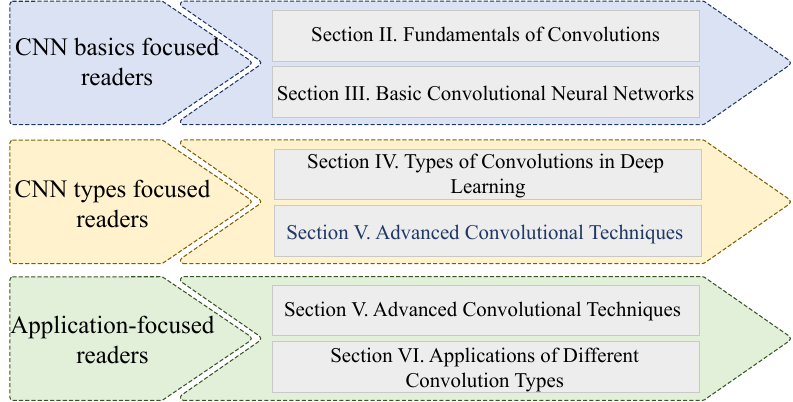}
    \caption{A text-based visual reading map that helps individuals navigate and comprehend the paper}
    \label{fig:readmap}
\end{figure}

The rest of our review paper follows (See Fig. \ref{fig:structure}): Section 2 of the paper will delve into the fundamentals of convolutions, elucidating their mathematical formulation, operational mechanics, and the role they play in the architecture of neural networks. Section 3 describes the basic parts of CNNs. In Section 4, The exploration will cover 2D convolutions, 1D convolutions for sequential data, and 3D convolutions for volumetric data. Section 5 of the research paper will investigate advanced convolutional techniques that have emerged in recent years. This will encompass topics such as transposed convolutions for upsampling, depthwise separable convolutions for efficiency, spatial pyramid pooling, and attention mechanisms within convolutions. Section 6 of the paper will highlight the real-world applications of different convolution types, showcasing their utility in image recognition, object detection, NLP, audio processing, and medical image analysis. In section 7 we discuss future trends and some open questions about CNNs. Section 8 is about the performance consideration of CNNs. In Section 9, we are going to talk about the platforms that are mostly used by researchers and developers, and in Section 10 about research fields that are popular or trending, then we have discussion in Section 11. By the end of this research in Section 8, readers will gain a profound understanding of the importance of convolutions in DL and 
Fig. \ref{fig:readmap} represents a reader map to visualize the flow of information within a text. It shows the connections between various sections, assisting readers in comprehending the overall structure of their preferred section following their needs.

\section{Fundamentals of Convolutions} 
\label{sec:Fundamentals} 
Convolutions form the foundation of crucial mathematical operations used to process data structured in grids, such as images, videos, and time series data \cite{ref26}. Originally used in signal processing, convolutions were used for analyzing and manipulating signals \cite{ref27}. In deep learning, convolutions serve as powerful feature extractors, enabling neural networks to efficiently learn from raw data \cite{ref26}\cite{ref27}. The essence of a convolution involves the sliding of a small filter, commonly known as a kernel, over the input data. At each position of this sliding operation, the kernel performs element-wise multiplication with the corresponding input values \cite{ref28}. Through this process, local patterns and relationships within the data are captured, enabling the model to acquire essential features like edges, textures, and shapes.

\subsection{Mathematical Formulation of Convolutions }
Mathematically, a 2D convolution between an input matrix (often representing an image) and a kernel can be represented as follows: 
\begin{equation}
\text{Output}(i,j) = \sum_{(x,y)} \text{Input}(x,y) \cdot \text{Kernel}(i-x, j-y)
\end{equation}

Here, Output denotes the resulting feature map, and Input represents the input matrix. The kernel, usually a small square matrix, defines the convolutional filter's weights. The convolution operation is performed by sliding the kernel over the input matrix, and at each position, the element-wise multiplication and summation are computed as described in the formula \cite{ref29}. For 1D convolutions, the mathematical formulation is similar, with the kernel sliding along a one-dimensional sequence, such as a time series or text data \cite{ref30}.

\subsection{Convolutional Operations in DL}
Convolutional operations form the core of CNNs, a highly prominent class of DL models widely utilized for various CV applications. Within a CNN, convolutions are typically integrated into specific layers referred to as convolutional layers \cite{ref31}. These layers are composed of multiple filters, each responsible for detecting distinct patterns in the input data \cite{ref139,ref140,ref141,ref142,ref143,ref144,ref145,ref146}. During the training phase, the model goes through the process of backpropagation and gradient descent to learn the optimal weights of the convolutional filters. This enables the model to automatically discern meaningful patterns within the data.
\begin{figure}[t]
    \centering
    \includegraphics[width=\columnwidth]{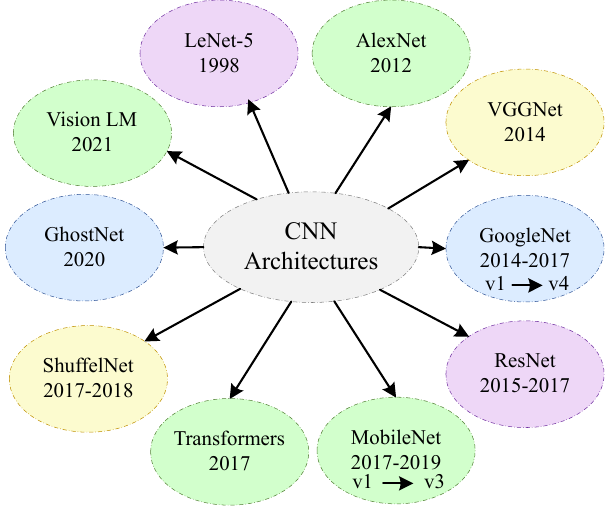}
    \caption{. A graphical representation of CNN architectures from 1998 to 2023}
    \label{fig:simple}
\end{figure}
Moreover, CNN architectures (See Fig. \ref{fig:simple} and Fig. \ref{fig:trend}) often incorporate pooling layers following the convolutional layers. As a result of pooling layers, feature maps generated by convolutions are downsampled, reducing computational complexity. Common pooling techniques include max-pooling and average pooling, which we will discuss about them in Section 3. B.
\begin{figure*}[t]
    \centering
    \includegraphics[width=\linewidth]{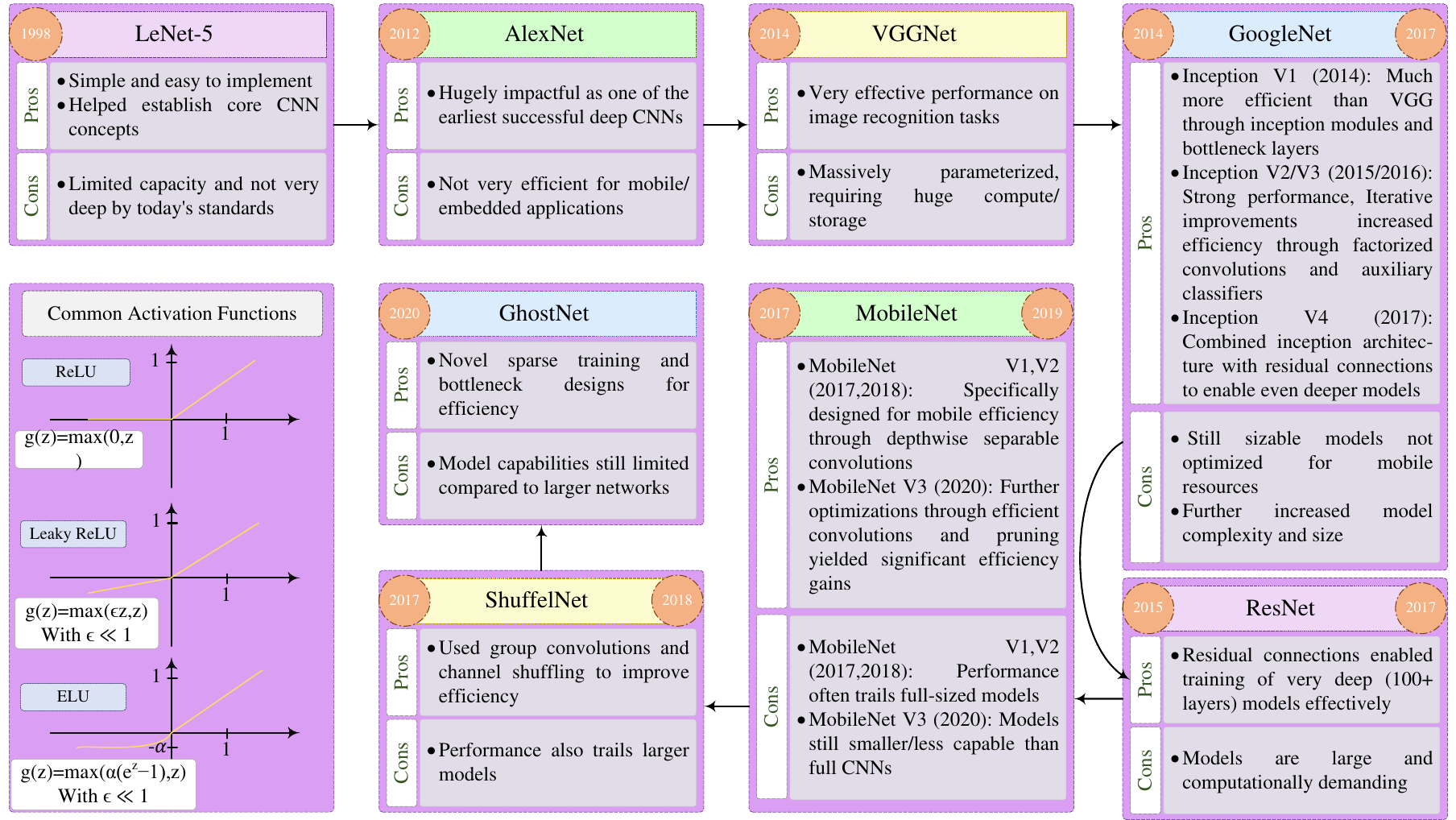}
    \caption{The flow of CNN architectures from 1998-2020 with their pros and cons represents that each CNN model is efficient for a specific application}
    \label{fig:trend}
\end{figure*}

\subsection{Wavelets}
Wavelets are an important mathematical tool that has numerous applications in fields such as signal processing and computer graphics. At their core, wavelets rely on convolution to analyze functions or continuous-time signals \cite{ref104}. By convolving the target function with wavelet basis functions at different scales, wavelets are capable of representing data with varying degrees of resolution \cite{ref109}.

Wavelet analysis uses small waves, called wavelets, as basis functions instead of the sine and cosine functions used in Fourier analysis \cite{ref105}. Wavelets have the advantage of analyzing properties of data locally in time and frequency instead of globally. This makes them well-suited for tasks such as edge detection, noise removal, and texture identification. The wavelet basis can also be adapted to the input signal or data being analyzed \cite{ref105}\cite{ref106}.

CNNs naturally lend themselves to wavelet analysis due to their intrinsic use of convolution operations \cite{ref107}\cite{ref108}. During training, the convolutional filters within CNNs can learn wavelet-like basis functions tailored to meaningfully represent the given input data distribution at multiple resolutions. By adopting the wavelet bases through gradient descent and backpropagation, CNNs gain an efficient multi-scale representation of patterns in the data \cite{ref108}\cite{ref109}.

A key characteristic of wavelets is their ability to decompose a signal into different frequency components, with high frequencies corresponding to detailed information and low frequencies corresponding to overall trends \cite{ref108}. A single-level wavelet decomposition breaks down the original signal into approximation and detail coefficients. The approximation contains lower frequency information, while the detail contains higher frequency or detailed information \cite{ref109}.

CNNs can utilize this multi-resolution decomposition property of wavelets by using convolutions to learn wavelet filters at each level \cite{ref108,ref109,ref110}. The output of each level becomes the input to the next, with the filters extracting more detailed features at higher levels after the removal of coarse information. This convolutional learning of adapted wavelet bases enables CNNs to hierarchically capture patterns across different scales for improved data representation \cite{ref110}.

In various image processing and computer vision tasks, the use of convolutional wavelets within CNNs has shown promising results. For applications like denoising, super-resolution, and texture synthesis, CNNs equipped with learned wavelet filters have achieved state-of-the-art performance by effectively representing key multi-scale characteristics of visual data \cite{ref110,ref111,ref112,ref113}. Convolutional wavelets also benefit segmentation, detection, and classification when combined with traditional convolutional filters within CNNs \cite{ref109}. In summary, wavelets provide a powerful tool for multi-scale analysis that CNNs can leverage through their inherent ability to learn localized basis functions via convolution operations.

\section{Basic Convolutional Neural Networks}
\label{sec:Basic}
\begin{figure}[t]
    \centering
    \includegraphics[width=\columnwidth]{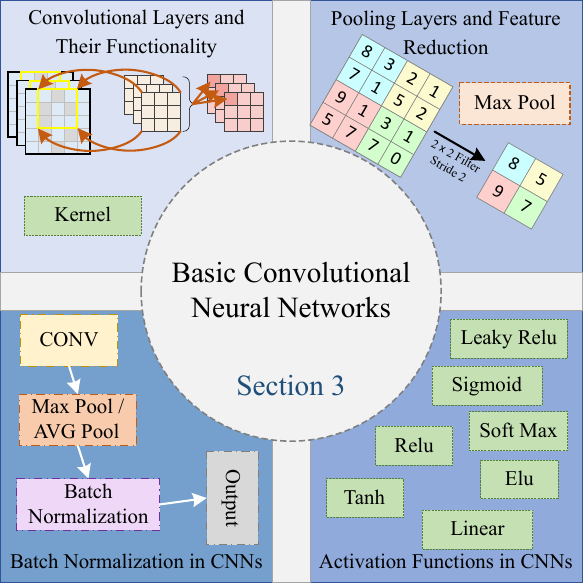}
    \caption{A graphical representation of Section 3}
    \label{fig:section 3}
\end{figure}
The CNN architecture typically consists of an initial input layer, followed by several critical components, including convolutional layers, pooling layers, and fully connected layers. This organized structure allows for the systematic processing of raw data, such as images, through a series of layers, which in turn enables the extraction of relevant features and facilitates making predictions.

The convolutional layers hold a central position in this architecture, as they employ learnable filters to process the input data. This operation is instrumental in detecting diverse patterns and features, thereby enhancing the network's ability to understand the underlying data. Following the convolutional layers, the pooling layers come into play, downsampling the output from the previous layers. This downsampling process reduces the spatial dimensions while retaining crucial information. By focusing on the most significant details, these layers contribute to translational invariance, a valuable aspect in applications like image recognition where object positions may vary. 

In Table \ref{tab:Basic Convolutional Neural Networks}, a comprehensive overview of the core components of basic CNNs is presented (also See Fig. \ref{fig:section 3}), encompassing convolutional layers, pooling layers, and activation functions. The table provides insights into their individual purposes, functionalities, dependencies on input size, parameters, feature maps, translational invariance, computational efficiency, output size, roles in the CNN architecture, and impact on model performance. Analyzing these aspects provides profound insights into the elements that contribute to the effectiveness and performance of CNNs, making it a valuable reference for researchers and practitioners in the field.

\subsection{Background of Deep Learning}

Deep learning, a prominent form of machine learning, encompasses the use of neural networks composed of multiple layers to acquire hierarchical representations of data \cite{ref17}. Taking inspiration from the intricate workings of the human brain, where neurons engage in processing and transmitting information to forge elaborate depictions of the world, DL models, also known as deep neural networks, showcase remarkable prowess in assimilating hierarchical features from raw data. This exceptional ability enables them to discern intricate patterns and achieve remarkable precision in predictions \cite{ref18}.

\begin{table*}[t]
\setlength\extrarowheight{3pt}
\centering
\caption{The Different Aspects of the Basic Convolutional Neural Networks}
\label{tab:Basic Convolutional Neural Networks}
\resizebox{\textwidth}{!}{%
\begin{tabular}{ccccc}
\hline
\rowcolor[HTML]{C0C0C0} 
\cellcolor[HTML]{C0C0C0}{\color[HTML]{000000} \textbf{Aspect}} &
  {\color[HTML]{000000} \textbf{Convolutional   Layers}} &
  \textbf{Pooling Layers} &
  \textbf{Activation   Functions} &
  \textbf{Batch Normalization} \\ \hline
\textbf{Purpose} &
  Feature extraction &
  Feature reduction &
  Introduce non-linearity &
  Training stabilization \\ \hdashline
\textbf{Functionality} &
  Detect patterns and textures &
  Downsample feature maps &
  Add non-linearity &
  Normalizing activations \\ \hdashline
\textbf{Input size dependency} &
  Depends on input dimensions &
  Reduces spatial dimensions &
  Independent of input &
  Depends on input size \\ \hdashline
\textbf{Parameters} &
  Learnable weights (kernels) &
  No parameters &
  No parameters &
  \begin{tabular}[c]{@{}c@{}}Learnable scaling \& \\ shifting parameters\end{tabular}   \\ \hdashline
\textbf{Feature maps} &
  Produce feature maps &
  No feature maps &
  No feature maps &
  No feature maps \\ \hdashline
\textbf{Translational invariance} &
  Not inherently invariant &
  Introduces some invariance &
  Independent of input &
  No Translational invariance \\ \hdashline
\textbf{Computational efficiency} &
  Computationally intensive &
  \begin{tabular}[c]{@{}c@{}}Reduces computation \\ complexity\end{tabular}  &
  Low computation cost &
  Enhanced training stability \\ \hdashline
\textbf{Output size} &
  \begin{tabular}[c]{@{}c@{}}May or may not \\ match the input size\end{tabular} &
  Reduced size &
  Unchanged &
  Unchanged \\ \hdashline
\begin{tabular}[c]{@{}c@{}}\textbf{Role in CNN} \\ \textbf{architecture}\end{tabular} &
  Central component &
  \begin{tabular}[c]{@{}c@{}}Interposed between\\ convolutions\end{tabular}   &
  \begin{tabular}[c]{@{}c@{}}Enable learning \\ complex relationships\end{tabular}  &
  \begin{tabular}[c]{@{}c@{}}Improve convergence,\\ ease of tuning\end{tabular}  \\ \hdashline
\begin{tabular}[c]{@{}c@{}}\textbf{Influence on} \\ \textbf{model performance}\end{tabular} &
  \begin{tabular}[c]{@{}c@{}}Significantly \\ impacts performance\end{tabular}  &
  \begin{tabular}[c]{@{}c@{}}Affects model \\ efficiency\end{tabular}  &
  \begin{tabular}[c]{@{}c@{}}Crucial for \\ Learning\end{tabular}  &
  \begin{tabular}[c]{@{}c@{}}Significantly \\impacts performance\end{tabular}  \\ \hdashline
\textbf{Interpretability} &
  Low &
  Low &
  Low &
  Normal \\ \hdashline
\textbf{Training complexity} &
  High &
  Low &
  Low &
  Normal \\\hdashline
\textbf{Memory usage} &
  Normal &
  Low &
  Low &
  Normal \\ \hline
\end{tabular}%
}
\end{table*}

The roots of DL can be traced back to the nascent endeavors surrounding artificial neural networks in the 1940s. However, the true resurgence and substantial remarkable materialized in the 1980s and 1990s, paving the way for its remarkable revival in the 21 century \cite{ref19}. Key catalysts driving this resurgence were the strides made in computational power, the vast availability of datasets, and the advent of efficient training algorithms, most notably backpropagation, which played a pivotal role \cite{ref20}. By harnessing these advancements, DL models attained the ability to process and analyze vast repositories of data, thus acquiring an aptitude for deciphering intricate patterns and making precise predictions.

The convergence of powerful hardware and sophisticated algorithms ushered in an era of remarkable accomplishments across diverse domains. Computer Vision (CV), natural language processing (NLP), and speech recognition (SR), among others, have witnessed remarkable strides through the transformative power of DL \cite{ref73}. This dynamic discipline's capacity to overcome more difficult problems and promote innovation across various industries is becoming more and more clear as it develops and advances.

\subsection{Introduction to Convolutional Neural Networks}
CNNs, an influential category of DL models, have emerged as a preeminent and extensively utilized algorithm within the realm of DL \cite{ref21}. Distinctive to CNNs is their capacity to engage in convolution calculations and operate proficiently on intricate structures. This characteristic has propelled CNNs to achieve remarkable breakthroughs in image analysis and feature extraction, bestowing upon them the ability to discern and efficiently classify features in images. Moreover, CNNs are renowned as shift-invariant artificial neural networks, a nomenclature that accentuates their capability to classify input information based on its hierarchical arrangement \cite{ref22}.

The hierarchical architecture of CNNs empowers them to process and extract features from input data in a shift-invariant manner \cite{ref22}. This implies that CNNs can adeptly recognize and classify objects within images, irrespective of their position or orientation. The realization of this shift-invariant attribute is accomplished through the application of convolutional layers, which employ filters in a sliding window fashion. These filters acquire the ability to detect specific patterns or features at various spatial scales, thereby enabling the network to encapsulate both local and global information. Consequently, CNNs exhibit profound proficiency in extracting meaningful features from images, facilitating a wide array of applications encompassing object detection, image recognition, and even image generation \cite{ref74}.

\subsection{Convolutional Layers and Their Functionality} 
Each convolutional layer comprises multiple filters, also referred to as kernels, which are small windows that slide over the input data \cite{ref32}. During the training phase, the weights of these filters are learned, and they function as feature extractors, identifying specific patterns, edges, and textures present in the input \cite{ref33}. When the filters move across the input, they create feature maps that emphasize important parts of the data as region of interest (ROI). These maps show where specific patterns in the input become active, helping the CNN recognize significant features crucial for later tasks like classification or detection \cite{ref34}. 

For example, in a CNN trained to identify cats in images, the filters may learn to recognize the patterns of fur, whiskers, and ears. As the filters convolve across an image of a cat, they generate feature maps that highlight these specific regions of interest. These feature maps indicate the activation of these cat-specific patterns and aid in accurately classifying the image as containing a cat.

\subsection{Pooling Layers and Feature Reduction }
Pooling layers are incorporated following convolutional layers to decrease the spatial dimensions of the feature maps, thereby reducing the computational complexity of the network \cite{ref35}. The most frequently utilized pooling techniques in CNNs are max-pooling and average-pooling \cite{ref37}. 

Max-pooling entails selecting the maximum value from a small region of the feature map, while average-pooling computes the average value. Pooling offers two primary advantages: first, it effectively reduces the number of parameters in the network, resulting in improved computational efficiency. Second, it introduces a level of translational invariance, signifying that minor spatial translations in the input data do not substantially impact the pooled outputs. This property enhances the CNN's ability to generalize better to variations in the input data.

For example, in image classification applications, after several convolutional and activation layers, a pooling layer can be used to downsample the feature map. This downsampling reduces the spatial resolution of the features, making it more computationally efficient to process and reducing the risk of overfitting. Additionally, because pooling computes either maximum or average values, it can capture the dominant features in an image regardless of their exact location, making the network more robust to slight variations in object position or orientation.
 
\subsection{Activation Functions in CNNs }
Activation functions play a vital role in CNNs as they are applied to the output of each neuron, introducing nonlinearity to the network and facilitating the learning of complex relationships between input data and their corresponding features. Within CNNs, several commonly used activation functions include Rectified Linear Units (ReLU) \cite{ref36}, which set negative values to zero while preserving positive values unchanged. Variants like Leaky ReLU \cite{ref36} and Parametric ReLU \cite{ref39} are also widely employed. The selection of the activation function is of great importance as it directly impacts the network's capacity to learn and make accurate predictions. By introducing nonlinearity, activation functions allow CNN to model intricate patterns and decision boundaries, thereby enhancing its performance across a range of tasks.

For example, in image classification applications, the ReLU activation function has been shown to effectively remove negative pixel values and emphasize positive pixel values, allowing CNN to identify important features and learn discriminative patterns. This enables the CNN to accurately classify different objects in images, such as correctly identifying whether an image contains a cat or a dog.

\subsection{Batch Normalization in CNNs }
Batch Normalization is a technique that helps stabilize and accelerate the training of CNNs \cite{ref78}. It normalizes the activations of each layer by centering and scaling the values using the mean and variance of each mini-batch during training. This process reduces internal covariate shifts, making the optimization process smoother and enabling the use of higher learning rates.

By normalizing activations, Batch Normalization allows for more aggressive learning rates, which leads to faster convergence and improved model generalization. Additionally, it acts as a regularizer, reducing the need for other regularization techniques like dropout.

Overall, Batch Normalization has become a standard component in CNN architectures, contributing to faster training, improved model performance, and increased ease of hyperparameter tuning. Its widespread adoption has significantly contributed to the success of modern CNNs in various CV and NLP applications.
For example, in image classification applications, Batch Normalization helps reduce overfitting by normalizing the input for each mini-batch during training. This ensures that the network learns robust features and avoids relying on specific pixel values or noise in the input data. As a result, the model becomes more generalized and performs better on unseen data.

\section{Types of Convolution in Deep Learning}
\label{sec:Types}
\begin{figure}[t]
    \centering
    \includegraphics[width=\columnwidth]{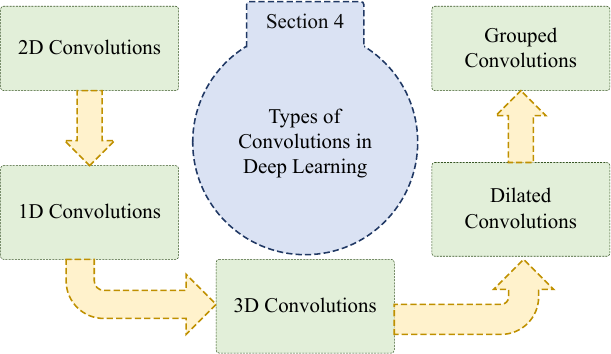}
    \caption{An overview of Section 4 structure}
    \label{fig:Section 4}
\end{figure}

In this section, our goal is to comprehensively explore the different convolution methods (See Fig. \ref{fig:Section 4}) commonly used in deep learning models. Table \ref{tab:Characteristics} presents a condensed overview of these convolution types, providing important information such as input data type, dimensionality, receptive field, computational cost, primary use case, memory consumption, parallelization capability, consideration of temporal information, and computational efficiency.
\begin{figure}[b]
    \centering
    \includegraphics[width=\columnwidth]{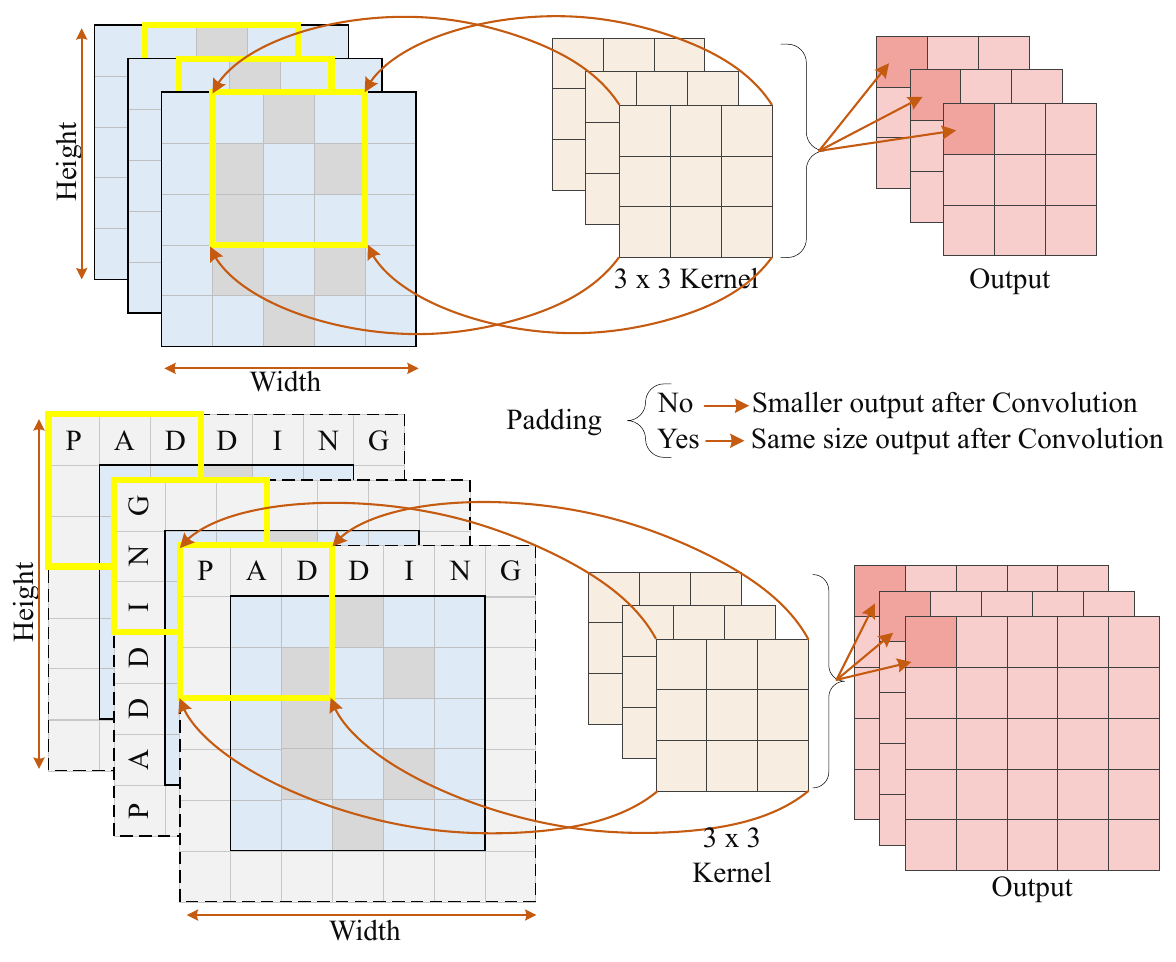}
    \caption{The Basic structure of CNN. a) represents CNN without Padding which causes the output image to become smaller. b) represents CNN without Padding which the output image is the same size as the input image}
    \label{fig:CNN 2dv2}
\end{figure} 

It is important to highlight that selecting the appropriate convolutional type relies on the particular task and dataset under consideration. For instance, when working with diverse data types, such as images or text, it may be necessary to employ distinct convolutional types to effectively capture relevant features. Moreover, considering the computational efficiency of each convolutional type becomes important for real-time applications or settings with limited resources. 

\begin{table*}[]
\centering
\setlength\extrarowheight{3pt}
\caption{The Comparison Provides an Overview of The Characteristics and Functionalities of Different Convolution Types}
\label{Characteristics}
\resizebox{\textwidth}{!}{%
\begin{tabular}{cccccc}
\hline
\rowcolor[HTML]{C0C0C0} 
\textbf{Convolution Type} &
  \textbf{2D Convolutions} &
  \textbf{1D Convolutions} &
  \textbf{3D Convolutions} &
  \textbf{Dilated Convolutions} &
  \textbf{Grouped Convolutions} \\ \hline
\textbf{Input Data Type} &
  Images &
  \begin{tabular}[c]{@{}c@{}}Sequential Data  \\ (e.g., Text)\end{tabular} &
  \begin{tabular}[c]{@{}c@{}}Volumetric Data \\ (e.g., Videos)\end{tabular} &
  Images &
  Images \\ \hdashline
\textbf{Dimensionality}           & 2D      & 1D      & 3D         & 1D, 2D  & 2D    \\ \hdashline
\textbf{Receptive Field}          & Local   & Local   & Volumetric & Local   & Local \\ \hdashline
\textbf{Computational Cost}       & Medium  & Low     & High       & Low     & High  \\ \hdashline
\textbf{Main Use Case} &
\begin{tabular}[c]{@{}c@{}}Image recognition, \\ Object detection\end{tabular}  & 
\begin{tabular}[c]{@{}c@{}}Text classification, \\ Sentiment analysis\end{tabular}  &
\begin{tabular}[c]{@{}c@{}}Semantic segmentation,\\ 3D medical imaging\end{tabular}  &
\begin{tabular}[c]{@{}c@{}}Image Filtering, \\ Image generation\end{tabular} & 
\begin{tabular}[c]{@{}c@{}}Large-scale \\ CNN architectures\end{tabular} \\ \hdashline
\textbf{Memory Consumption}       & Medium  & Low     & High       & Low     & Low   \\ \hdashline
\textbf{Parallelization}          & Limited & Limited & Limited    & Limited & High  \\ \hdashline
\begin{tabular}[c]{@{}c@{}}\textbf{Use of Temporal}  \\  \textbf{Information}\end{tabular} &
  Not applicable &
  \begin{tabular}[c]{@{}c@{}}Captures temporal \\ patterns\end{tabular}  &
  \begin{tabular}[c]{@{}c@{}}Captures spatial \\ temporal patterns\end{tabular}  &
  Not applicable &
  Not applicable \\ \hdashline
\begin{tabular}[c]{@{}c@{}}\textbf{Computational} \\ \textbf{Efficiency}\end{tabular}\ & Medium  & High    & Medium     & High    & High  \\ \hline
\end{tabular}%
}
\end{table*}

\subsection{2D Convolutions}

2D convolutions (See Fig. \ref{fig:CNN 2dv2}) serve as the foundational elements in CNNs, particularly for applications related to CV. They are predominantly utilized for processing two-dimensional data, such as images, which can be represented as a grid of pixels. During this convolutional operation, a 2D kernel slides over the input image, enabling the capture of local patterns and the extraction of relevant features \cite{ref27}. The primary application of 2D convolutions lies in image recognition, wherein the model learns to identify essential patterns, including edges, textures, and object components, thereby facilitating high-level recognition applications \cite{ref40}. 

2D convolutions have found use in a variety of fields, including signal processing, CV, and NLP in addition to image recognition. CNNs have completely changed CV processes like object detection, image segmentation, and facial recognition. CNNs can more accurately and efficiently analyze the spatial relationships and hierarchical structures present in images by using 2D convolutions. When learned filters slide across the input image, a CNN can learn to find and locate different objects in images, such as in object detection tasks. This helps the network accurately detect objects even in complicated scenes, as it can identify important patterns of various sizes.

Moreover, CNNs can also be learned to categorize and compare faces by analyzing facial features using 2D convolutions in facial recognition. This makes it possible to create systems like access control and identity verification.

\subsection{1D Convolutions for Sequential Data }
\begin{figure}[b]
    \centering
    \includegraphics[width=\columnwidth]{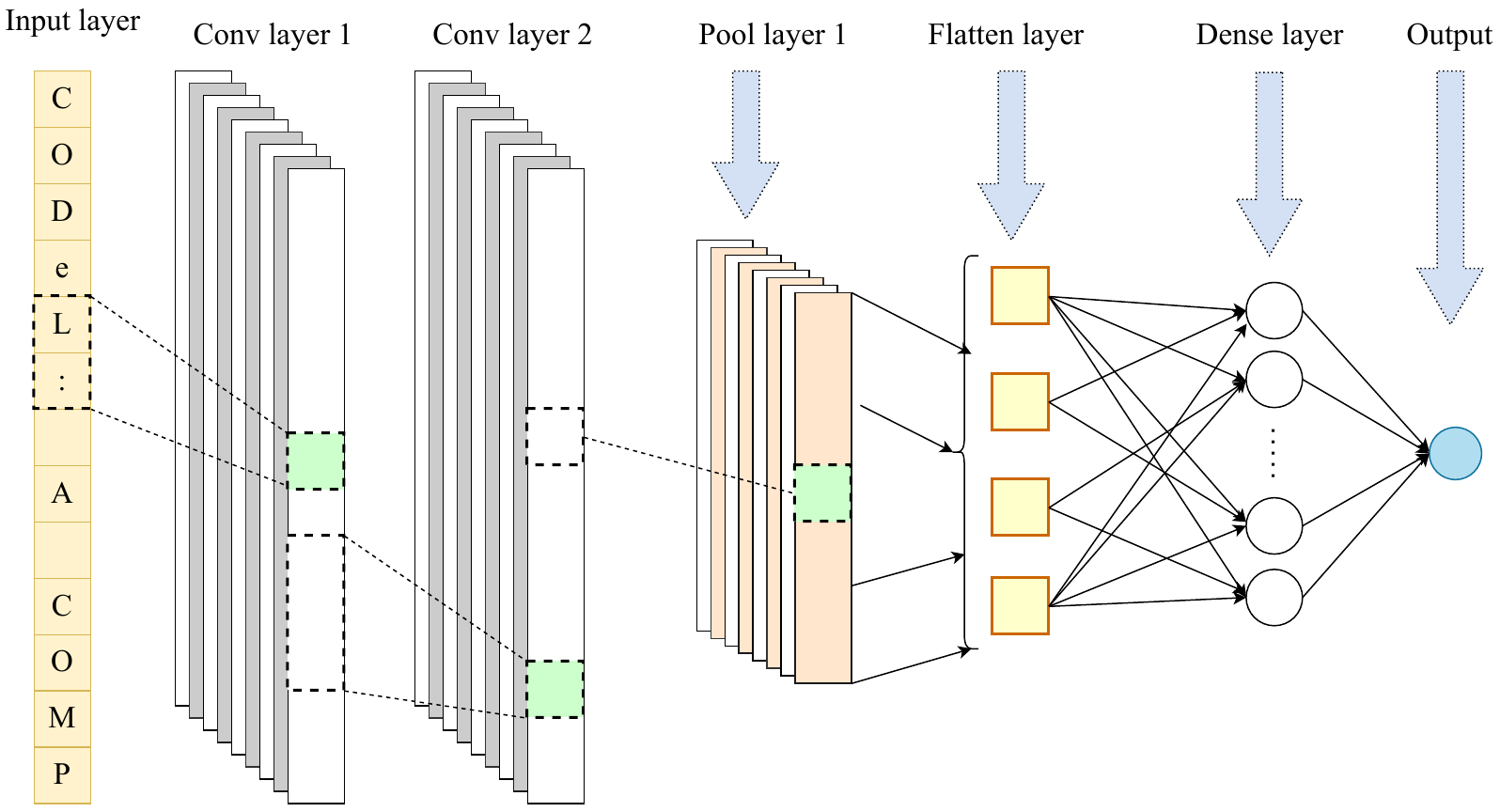}
    \caption{An overview to simple one-dimensional (1D) Convolution Neural Network with Two Convolution layer}
    \label{fig:1d cnn}
\end{figure}

One-dimensional (1D) convolutions (See Fig. \ref{fig:1d cnn}) are specially designed for working with sequential data like time series, audio signals, and natural language. Unlike their two-dimensional counterparts, 1D convolutions operate on a single line, allowing them to detect patterns that develop over time \cite{ref41}. In the field of natural language processing, 1D convolutions are widely used in tasks such as classifying text and analyzing sentiments. They help the model identify complex patterns in sequences of words and understand how these words are related to each other \cite{ref42}. 
1D convolutions have also been successfully applied to audio signal processing applications such as SR and music analysis. By analyzing the temporal patterns of audio signals, these models can extract meaningful features that capture the underlying structure and characteristics of the sound. This has proven to be particularly useful in applications like speaker identification and emotion recognition, where the sequential nature of the audio data is sequential. 

For example, in speaker identification, 1D convolution can analyze the sequential patterns of an individual's voice and learn to associate certain patterns with specific speakers. This allows the model to accurately identify and differentiate between different speakers in an audio recording. In emotion recognition, 1D convolutions can analyze the temporal changes in pitch, tone, and intensity of an audio signal to classify the emotional state of the speaker, such as happiness, sadness, or anger. This helps in detecting and understanding the underlying emotions conveyed through speech, which can be useful in various applications like customer sentiment analysis, virtual assistants, and mental health monitoring.

\subsection{3D Convolutions for Volumetric Data }
Three-dimensional (3D) convolutions are specifically designed to handle volumetric data, such as 3D medical images or video data \cite{ref43}. 3D convolutions possess the capability to simultaneously process spatial and temporal dimensions, thereby capturing intricate patterns and distinctive features across all three dimensions. In medical imaging, 3D convolutions are vital in jobs like finding where tumors are. The model uses 3D medical scans to figure out where the important spatial and surrounding details are, which helps accurately locate and describe tumors \cite{ref44}\cite{ref45}.

The use of 3D convolutions has gone beyond just tumors and is used in various medical imaging tasks like picking out different parts of the body, spotting issues, and classifying diseases. This method lets the model see the whole volume of a medical scan, rather than just individual parts, and consider how different slices are related in space. This comprehensive approach allows the model to effectively capture the overall structure of the target organ or an anomaly, resulting in improved diagnostic accuracy and better patient outcomes. 

For instance, in tumor segmentation, 3D convolutions can be used to analyze a series of consecutive medical scans to identify the size and location of tumors over time, allowing doctors to track their growth and plan targeted treatments. This helps improve the accuracy and efficiency of tumor identification, leading to better patient outcomes.

In addition to operating on raw medical images and videos, 3D convolutions can be applied to process point cloud data through voxelization \cite{ref101}. As point clouds represent 3D geometry as an unordered set of points without connectivity, a common approach is to first discretize the continuous 3D space into regular volumetric grids called voxels. Each voxel is assigned a feature vector, such as the number of points or aggregated point properties within its volume.

Voxelizing the point cloud allows existing 3D convolutional kernel operations to be directly applied. Early works divided the spatial domain into coarse voxels and maxpooled point features inside each voxel \cite{ref101}. More advanced methods utilize sparse convolutions over fine-grained voxels or use dilated kernels with gaps to control the receptive field size. Multi-scale voxels have also been explored to capture both local and global point features \cite{ref126}\cite{ref127}.

After 3D convolution and pooling, the extracted voxel features can be decoded back to the original point cloud domain for subsequent 3D fully connected or Transformer layers \cite{ref130}. Voxel representation serves as an efficient intermediary that not only maintains the spatial structure required by CNNs but also allows points of variable density\cite{ref128}\cite{ref129}\cite{ref130}. This two-stage voxel-based approach enables end-to-end training of 3D CNNs for point clouds.

\subsection{Dilated Convolutions and Their Advantages} 
\begin{figure}[b]
    \centering
    \includegraphics[width=\columnwidth]{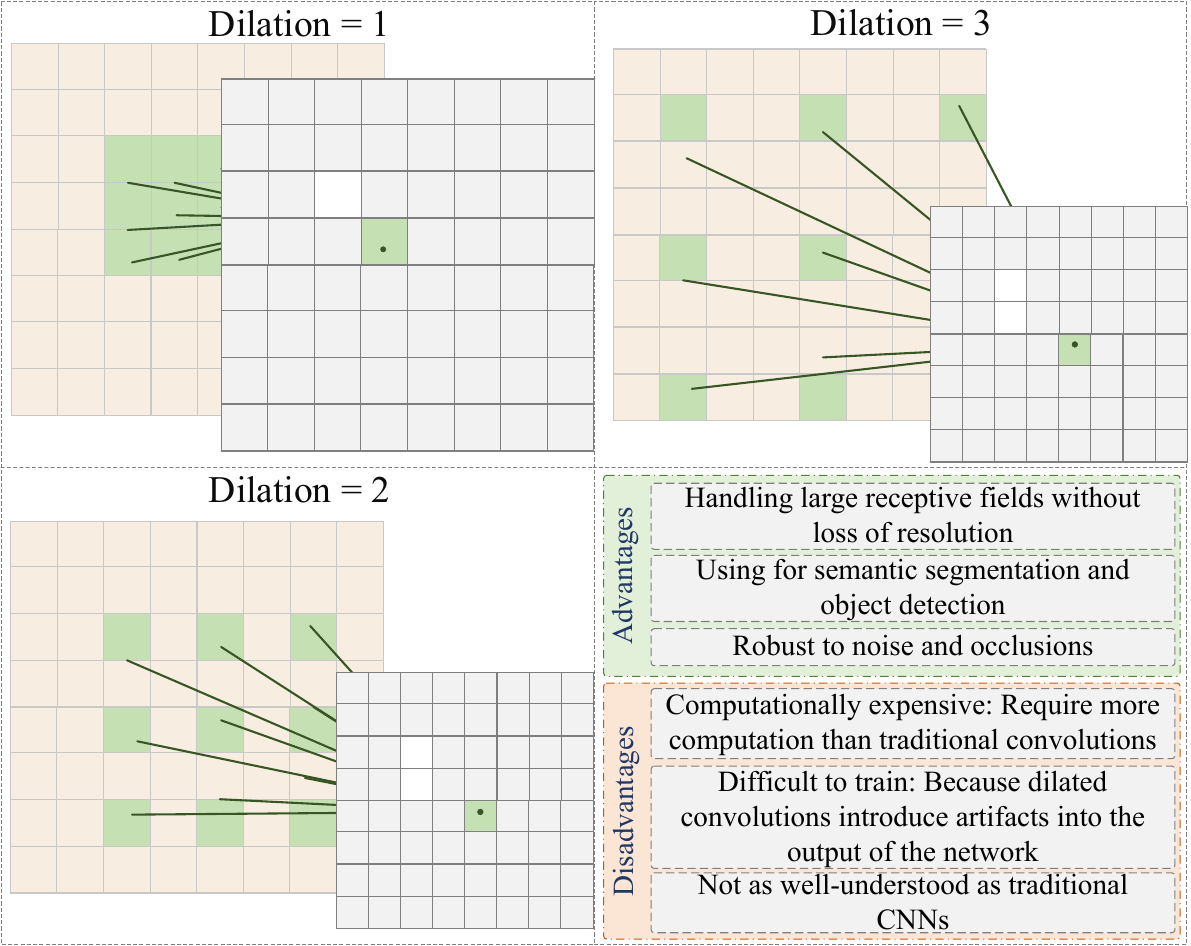}
    \caption{Dilation Convolution with multiple dilation rate with 3 x 3 kernel size \cite{ref74}}
    \label{fig:Dilation}
\end{figure}

Dilated convolutions (See Fig. \ref{fig:Dilation}), also known as atrous convolutions, are a variant of traditional convolutions that introduce gaps (dilation) between kernel elements. This gap enables for an increased receptive field without increasing the number of parameters, making dilated convolutions more computationally efficient \cite{ref46}. 
Dilated convolutions find application in applications like semantic segmentation, where they enable the model to capture broader contextual information without compromising computational efficiency \cite{ref47}. 

In semantic segmentation applications, dilated convolutions are particularly useful because they enable the model to capture broader contextual information. By introducing gaps between kernel elements, dilated convolutions increase the receptive field without adding more parameters. This means that the model can understand the surrounding context of each pixel or object in the image without sacrificing computational efficiency. This value is important in applications like semantic segmentation, where accurately identifying and classifying objects within an image is essential. 

\subsection{Grouped Convolutions for Efficiency}

\begin{figure}[b]
    \centering
    \includegraphics[width=\columnwidth]{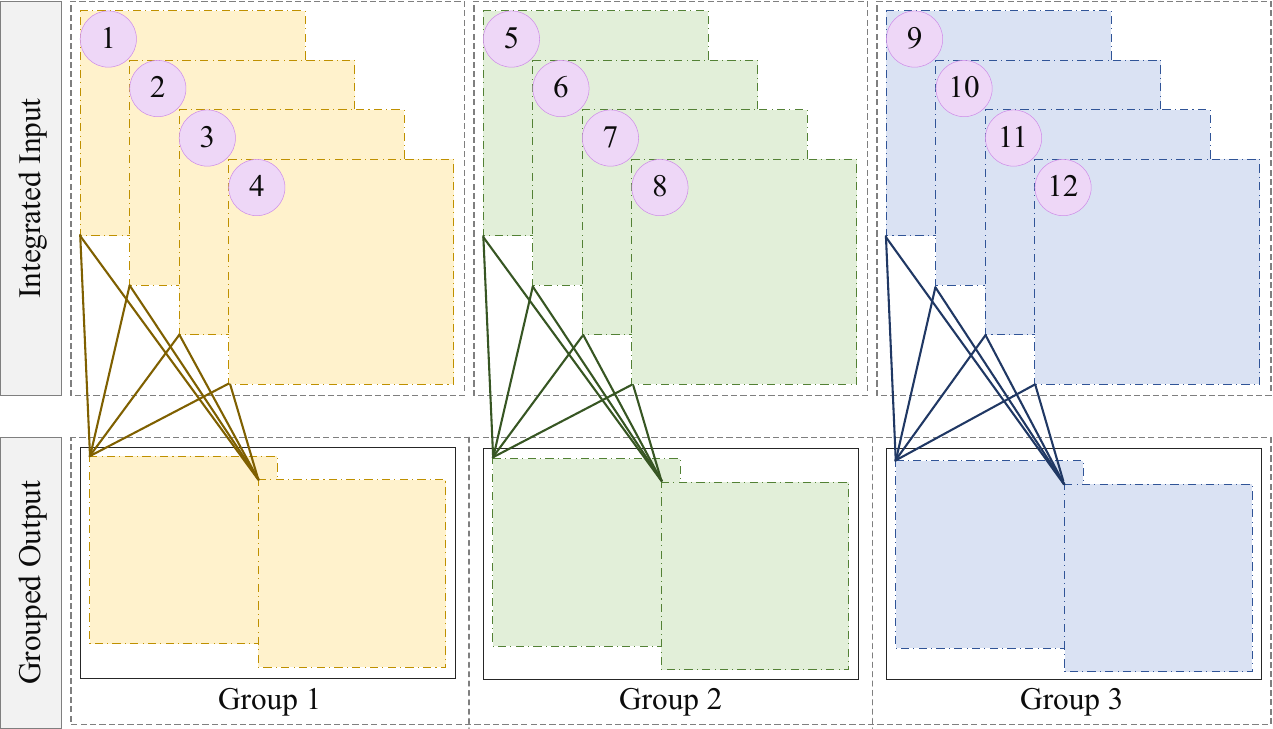}
    \caption{Grouped convolution involves dividing the channels of a convolutional layer into 3 groups}
    \label{fig:Grouped convolution}
\end{figure}

Grouped convolutions (See Fig. \ref{fig:Grouped convolution}) involve dividing the input and output channels of a convolutional layer into groups. Within each group, separate convolutions are performed, which are then concatenated to produce the final output. This technique significantly reduces computational cost and memory consumption while promoting model parallelism \cite{ref48}. Grouped convolutions are commonly used in large-scale CNN architectures to reduce training time and enhance the scalability of DL models \cite{ref49}.

In addition to reducing computational cost and memory consumption, grouped convolutions also offer other advantages. One of the main benefits is improved model parallelism, which provides for better utilization of parallel computing resources. This is especially important in large-scale CNN architectures where training time can be a bottleneck. By dividing the input and output channels into groups, the convolutions can be performed in parallel, speeding up the entire training process. Furthermore, the scalability of DL models is enhanced with grouped convolutions, making it easier to deal with larger datasets and more complex applications. 

\begin{figure*}[h]
    \centering
    \includegraphics[width=\linewidth]{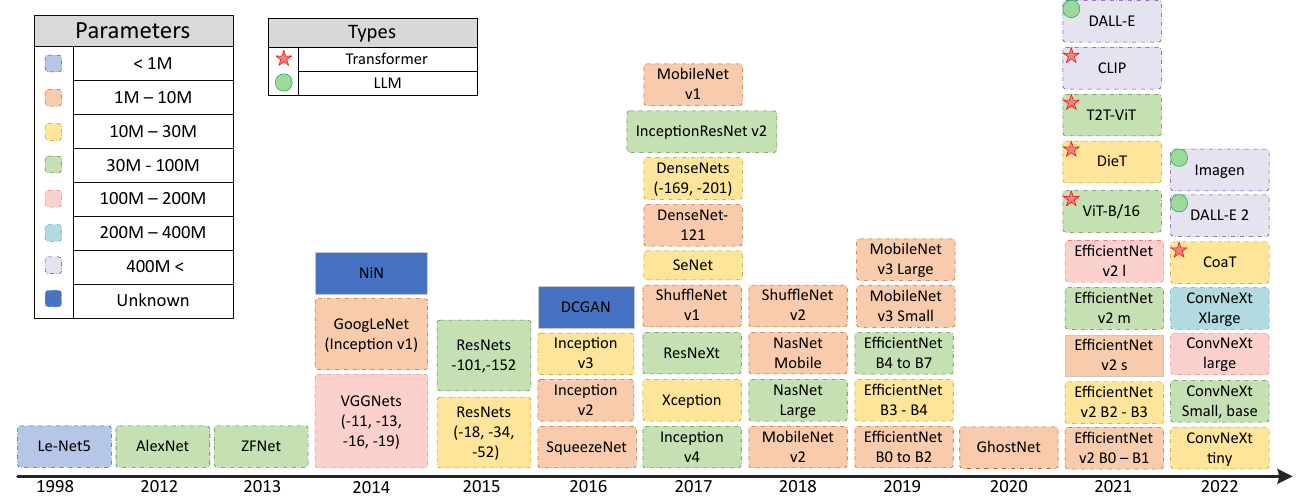}
    \caption{The detailed overview of advanced convolutions techniques}
    \label{fig:CNN trend}
\end{figure*}
For example, in image classification applications, a large-scale CNN architecture such as ResNet can benefit from model parallelism using grouped convolutions. By dividing the input and output channels into groups, different subsets of the model can be trained in parallel on multiple GPUs or distributed systems. This not only reduces the training time but also allows for better resource utilization, eventually improving the scalability of the DL model to handle larger datasets and more complex image recognition applications.

In conclusion, DL offers a diverse range of convolutional techniques to accommodate different data types and applications. From 2D convolutions for image recognition to 1D convolutions for sequential data and 3D convolutions for volumetric data, each convolution type has its unique advantages. Additionally, dilated convolutions and grouped convolutions serve as efficient alternatives, addressing specific challenges in DL models. Understanding the characteristics and applications of these convolution types empowers researchers and practitioners to design efficient and effective models for a wide array of applications.

\subsection{Evolution of CNN Architectures}
Since the early origins of CNNs, there has been a rapid evolution in CNN architectures (See Fig. \ref{fig:CNN trend}) \cite{ref49} over the past decade to enhance performance and efficiency [51]. Some key developments include:
\begin{itemize}
    \item Inception modules (2014) - The Inception architecture introduced convolutional blocks with multiple filter sizes to capture features at various scales \cite{ref52}. This improves both accuracy and computational efficiency.
    \item ResNets (2015) - Residual networks allow the training of much deeper CNNs through shortcut connections that bypass multiple layers \cite{ref53}. They reduce degradation in very deep models.
    \item DenseNets (2016) - These connect each layer to all subsequent layers for maximum information flow and feature reuse. This reduces the number of parameters \cite{ref54}.
    \item MobileNets (2017) - Designed specifically for mobile applications, they use depthwise separable convolutions to minimize model size and latency \cite{ref55}.
    \item EfficientNets (2019) - By systematically scaling network dimensions, these achieve much better efficiency-accuracy trade-offs \cite{ref55}.
\end{itemize}
The evolution of CNN architectures (See Fig. \ref{fig:CNN trend}) has been crucial to their widespread adoption across vision applications.

\section{Advanced Convolutional Techniques}
\label{sec:Advanced} 
\begin{figure}[t]
    \centering
    \includegraphics[width=\linewidth]{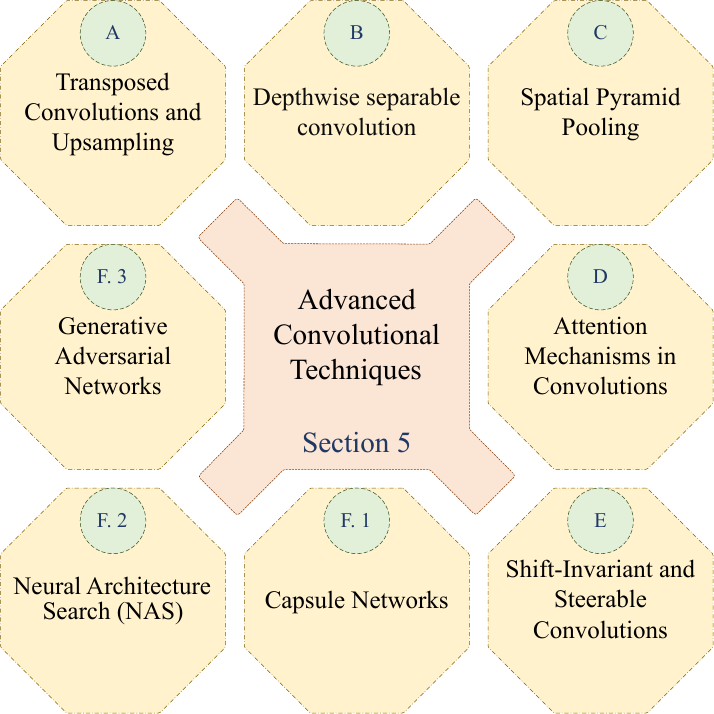}
    \caption{The trend of CNNs over time based on the released year and amount of parameters and their types}
    \label{fig:section5}
\end{figure}

This section provides a detailed overview of advanced convolutional techniques (See Fig. \ref{fig:section5}). A clear and informative summary of these techniques is available in Table \ref{tab:Advance-Part1}. By reviewing this table, readers can gain a better understanding of the state-of-the-art convolutional techniques and their potential uses. 

\subsection{Transposed Convolutions and Upsampling}
Transposed convolutions—also referred to as deconvolutions or fractionally stridden convolutions—are sophisticated methods for upsampling feature maps \cite{ref57}. Transposed convolutions, as opposed to conventional convolutions, increase the feature map size, enabling the model to reconstruct higher-resolution representations from lower-resolution inputs \cite{ref58}. Traditional convolutions reduce spatial dimensions. In processes like image segmentation \cite{ref59}, image creation \cite{ref60}, and image-to-image translation \cite{ref61}, they are essential. Transposed convolutions employ padding and stride values to regulate the upsampling process and learnable parameters to choose the output size.

Transposed convolution can create artifacts or checkerboard patterns in generated feature maps, due to overlapping receptive fields. To prevent this, stride, padding, and dilation are used to control the output resolution and reduce these artifacts. In the field of image generation, transposed convolutions are used to upscale low-resolution images into high-resolution ones. To ensure the generated images are free of artifacts or checkerboard patterns, stride, padding, and dilation are adjusted to control the output resolution and enhance the quality of the generated images.

\begin{table*}[!htb]
\setlength\extrarowheight{3pt}
\centering
\caption{The Comparison Provides an Overview of The Characteristics and Functionalities of Different Convolution Types - Part 1}
\label{tab:Advance-Part1}
\resizebox{\textwidth}{!}{%
\begin{tabular}{cccccc}
\hline
\rowcolor[HTML]{C0C0C0}
\textbf{Convolution Technique} &
\textbf{Transposed Convolutions} &
\textbf{DSC} &
\textbf{SPP} &
\textbf{Attention Mechanism} &
\textbf{Shift-Invariant} \\ \hline
\textbf{Purpose} &
Upsampling &
Parameter Reduction &
\begin{tabular}[c]{@{}c@{}}Handling Varying\\ Input Sizes\end{tabular}  &
\begin{tabular}[c]{@{}c@{}}Focus on Relevant\\ Features\end{tabular}  &
Invariance \\
\textbf{Parameters} &
Learnable &
Learnable &
No parameters &
Learnable &
Learnable \\
\textbf{Computational Cost} &
High &
Low &
Low &
Normal &
High \\
\textbf{Parameter Efficiency} &
Low &
High &
High &
Low &
Normal \\
\textbf{Upsampling} &
Yes &
No &
No &
No &
No \\
\textbf{Spatial Handling} &
Spatially Invariant &
Spatially Invariant &
Variable regions &
Spatially Invariant &
Spatially Invariant \\
\begin{tabular}[c]{@{}c@{}}\textbf{Long-range}\\ \textbf{Dependencies}\end{tabular} &
No &
No &
No &
Yes &
No \\
\textbf{Translation Invariance} &
Yes &
Yes &
Yes &
Yes &
Yes \\
\textbf{Rotation Invariance} &
No &
No &
No &
No &
No \\
\textbf{Interpretability} &
Low &
Low &
Low &
Low &
Low \\
\textbf{Model Size} &
Large &
Small &
Small &
Small &
Large \\
\textbf{Versatility} &
Normal &
High &
High &
High &
Normal \\ 
\textbf{Practical Applications} &
 \begin{tabular}[c]{@{}c@{}}Image Segmentation, \\ Image Super-Resolution,\\  Image Generation\end{tabular} &
  \begin{tabular}[c]{@{}c@{}}Mobile Vision Applications,\\  Real-time Object Detection\end{tabular} &
  \begin{tabular}[c]{@{}c@{}}Image Classification,\\  Object Detection,\\  Semantic Segmentation\end{tabular} &
  \begin{tabular}[c]{@{}c@{}}Image Captioning,\\ Visual Question Answering\end{tabular} &
  \begin{tabular}[c]{@{}c@{}}Image Recognition,\\  Object Detection,\\ Image Filtering\end{tabular}  \\ \hline

\end{tabular}%

}
\end{table*}

\begin{table*}[!htb]
\centering
\setlength\extrarowheight{3pt}
\caption{The Comparison Provides an Overview of The Characteristics and Functionalities of Different Convolution Types - Part 2}
\label{tab:Advance-Part2}
\resizebox{\textwidth}{!}{%
\begin{tabular}{ccccccc}
\hline
\rowcolor[HTML]{C0C0C0}
\textbf{Convolution Technique} &
\textbf{Steerable Convolution} &
\textbf{Capsule Networks} &
\textbf{NAS} &
\textbf{GAN} &
\textbf{VIT} \\ \hline
\textbf{Purpose} &
Efficiency and Invariance &
Invariance &
Efficiency &
Synthesis &
Long-range dependencies \\
\textbf{Parameters} &
Learnable &
Learnable capsules &
Architecture search &
Learnable &
Learnable \\
\textbf{Computational Cost} &
Low &
High &
High &
High &
Higher \\
\textbf{Parameter Efficiency} &
High &
Normal &
High &
Low &
Normal \\
\textbf{Upsampling} &
No &
No &
No &
No &
No \\
\textbf{Spatial Handling} &
Spatially Invariant &
Spatially Invariant &
Spatially variant &
Spatially Invariant &
Spatially Invariant \\
\textbf{Long-range Dependencies} &
No &
No &
No &
No &
Yes \\
\textbf{Translation Invariance} &
Yes &
Yes &
Yes &
No &
Yes \\
\textbf{Rotation Invariance} &
Yes &
Yes &
No &
No &
Yes \\
\textbf{Interpretability} &
Low &
Low &
Low &
Low &
High \\
\textbf{Model Size} &
Normal &
Normal &
Large &
Large &
Large \\
\textbf{Versatility} &
Low &
Low &
Low &
Low &
High \\
\textbf{Practical Applications} &
 \begin{tabular}[c]{@{}c@{}}Image Filtering, \\ Edge Detection, \\ Pattern Recognition\end{tabular} &
  \begin{tabular}[c]{@{}c@{}}Object Recognition, \\ Image Segmentation,\\ Medical Imaging\end{tabular} &
  \begin{tabular}[c]{@{}c@{}}Customized CNN Architectures,\\  Resource-Constrained Devices\end{tabular} &
  \begin{tabular}[c]{@{}c@{}}Image Synthesis, \\ Style Transfer,\\ Data Augmentation\end{tabular} &
  \begin{tabular}[c]{@{}c@{}}Image recognition, NLP,\\  diverse tasks\end{tabular} \\
\hline
\end{tabular}%
}
\end{table*}
\subsection{Depthwise Separable Convolutions (DSC)} 
\begin{figure}[b]
    \centering
    \includegraphics[width=\linewidth]{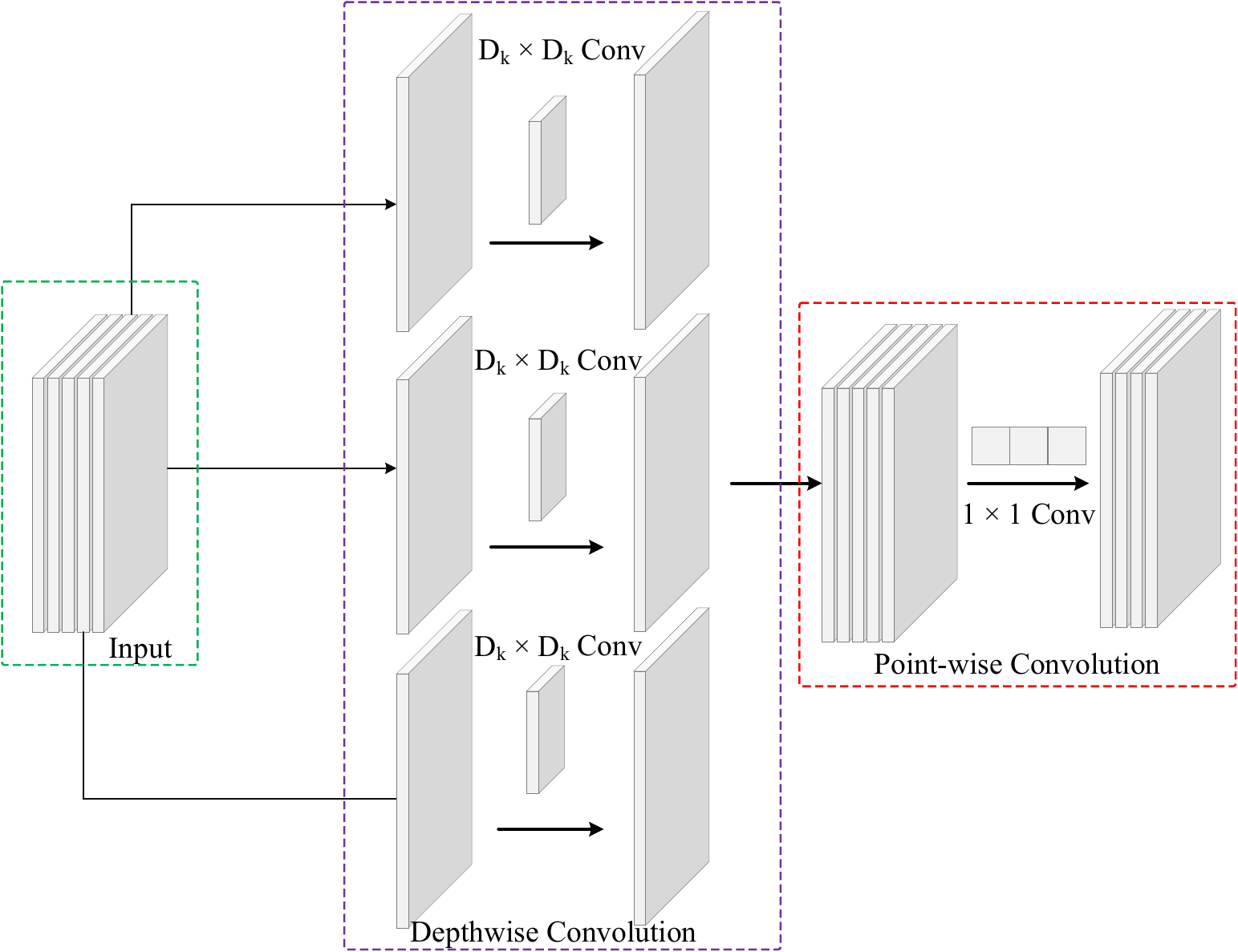}
    \caption{The Box with Purple color represents the Depthwise Convolution and the box with red color represents Pointwise Convolution (in pointwise a 1 x 1 convolution is used)}
    \label{fig:Depthwise}
\end{figure}

Depthwise separable convolutions (See the purple box in Fig. \ref{fig:Depthwise}) are an efficient alternative to traditional convolutions, particularly in resource-constrained environments \cite{ref62}\cite{ref63}. They split the convolution process into two steps (See Fig. \ref{fig:Depthwise}) depthwise convolutions \cite{ref64} and pointwise convolutions \cite{ref65} \cite{ref276,ref277,ref278,ref279}. Depthwise convolutions apply a separate kernel to each input channel, capturing spatial patterns independently for each channel. Pointwise convolutions then use 1x1 convolutions to combine the output channels from the depthwise step, effectively aggregating the information \cite{ref66}. Depthwise separable convolutions significantly reduce the number of parameters and computation while maintaining model performance, making them popular in mobile and embedded applications \cite{ref67}. 

By decoupling spatial filtering from cross-channel filtering, depthwise convolution achieves higher computational efficiency and is well-suited for resource-constrained environments. MobileNet and Xception are popular CNN architectures that use depthwise convolution to reduce model size and improve inference speed without compromising performance significantly.

\subsection{Spatial Pyramid Pooling (SPP) }
Spatial pyramid pooling (SPP) is a technique used to handle inputs of varying sizes and aspect ratios in CNNs \cite{ref68}\cite{ref280,ref281,ref282,ref283,ref284,ref285}. It divides the input feature maps into different regions of interest and applies max-pooling or average-pooling to each region independently. The resulting pooled features are then concatenated to form a fixed-length representation, which is fed into fully connected layers for further processing. SPP enables the CNN to accept input images of different sizes and produces consistent feature maps, making it useful in object detection and image segmentation applications \cite{ref69}. 

\subsection{Attention Mechanisms in Convolutions}
Attention mechanisms in convolutions allow the model to focus on relevant parts of the input, emphasizing specific regions during feature extraction \cite{ref70}. These mechanisms assign weights to different spatial locations based on their importance. Self-attention mechanisms \cite{ref70}, like those used in transformers, have been adapted for use in convolutions. They enable the network to capture long-range dependencies and context, improving the model's ability to recognize complex patterns and relationships. 

\subsection{Shift-Invariant and Steerable Convolutions }
Shift-invariant convolutions are designed to be insensitive to small translations in the input data \cite{ref71} \cite{ref286,ref287,ref288}. They ensure that the learned features remain consistent regardless of the object's position within the input image. This property is crucial for object detection applications, where the object's location might vary within the image \cite{ref27}. Steerable convolutions are filters that can be rotated to different angles, allowing the model to learn orientation-sensitive features in an orientation-invariant manner \cite{ref289,ref290,ref291}. These convolutions are often used in applications like text recognition, where the orientation of text can vary.

\subsection{Recent Advancements and Innovations}
\subsubsection{Capsule Networks}
Capsule Networks, introduced by Geoffrey Hinton and his team, is a revolutionary advancement in CNNs \cite{ref75}. They aim to address the limitations of traditional CNNs, particularly in handling spatial hierarchies and viewpoint variations \cite{ref292,ref293,ref294,ref295,ref296,ref297,ref298}. Capsule Networks use capsules as fundamental units, which are groups of neurons that represent various properties of an entity, such as its pose, deformation, and parts.

Capsule Networks offer dynamic routing mechanisms to route information between capsules, allowing them to model complex hierarchical relationships more effectively. This enables the network to recognize objects with various poses and appearances, making Capsule Networks more robust to transformations and occlusions. 

\subsubsection{Neural Architecture Search for Convolutions}
Neural Architecture Search (NAS) is an automated approach to designing CNN architectures \cite{ref76}\cite{ref81}. Instead of relying on human-designed architectures, NAS employs search algorithms and neural networks to discover architectures that perform well on specific applications \cite{ref76}. This technique has led to the development of state-of-the-art CNNs that outperform hand-crafted models \cite{ref299,ref300,ref301,ref302,ref303,ref304,ref305,ref306,ref307,ref308,ref309}.

NAS for convolutions involves exploring various convolutional designs, including different kernel sizes, depths, and connectivity patterns \cite{ref82}. It evaluates each architecture on a validation set, and through a process of evolution or optimization, identifies the best-performing architecture.

In the scenario of self-autonomous vehicle navigation, NAS for convolutions could be used to design an optimal convolutional neural network architecture specifically tailored for processing and analyzing various types of visual data collected by the vehicle's sensors. By exploring different convolutional designs, such as varying kernel sizes, depths, and connectivity patterns, NAS could identify the most effective architecture for accurately detecting objects and recognizing road signs in real-time. This would ultimately improve the vehicle's ability to navigate autonomously and make informed decisions based on its visual perception.

\subsubsection{Generative Adversarial Networks}
\begin{figure}[t]
    \centering
    \includegraphics[width=\linewidth]{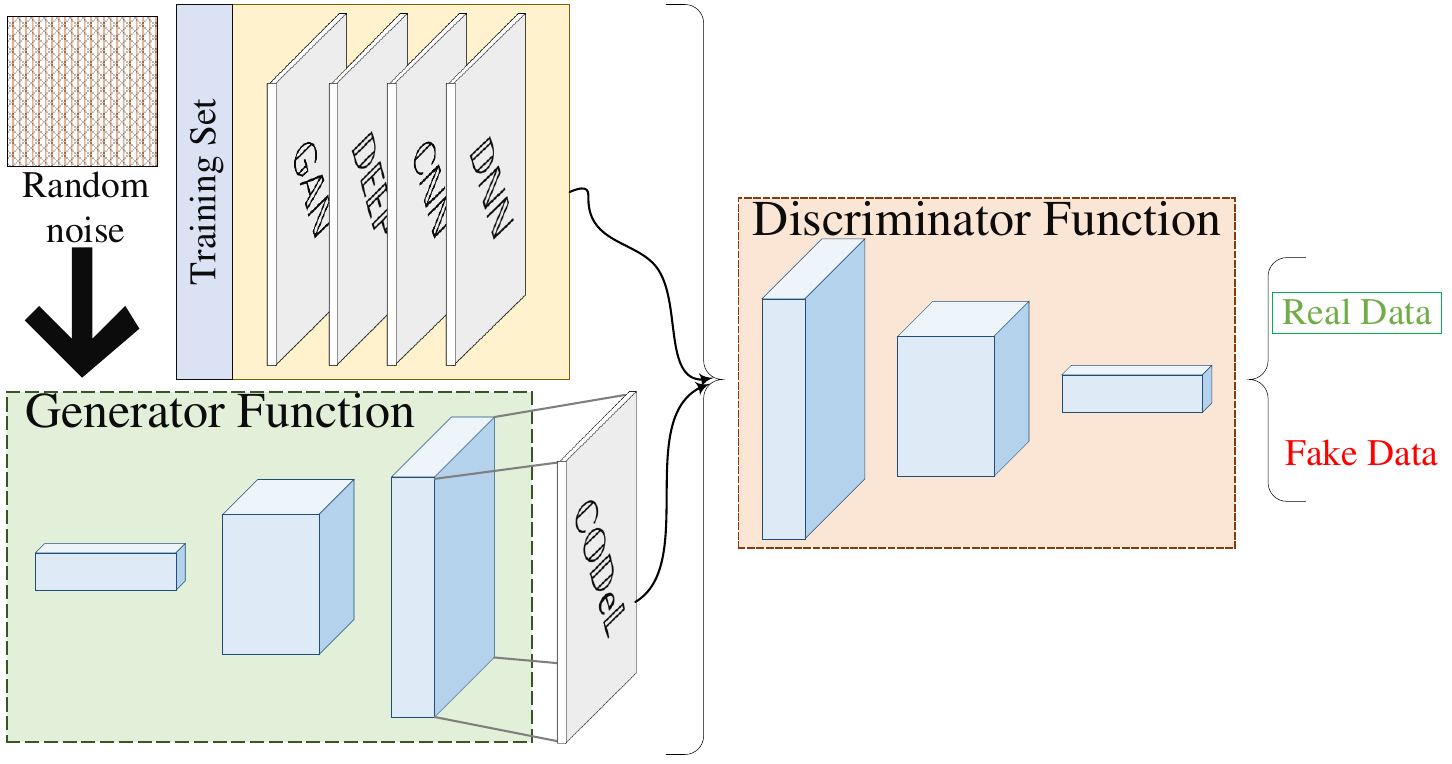}
    \caption{A simple GAN architecture represented to detect real and fake data which generator has generated}
    \label{fig:GAN}
\end{figure}

Generative Adversarial Networks (GANs) are a class of DL models used for generative applications, such as image synthesis, style transfer, and data augmentation \cite{ref310,ref311,ref312,ref313,ref314,ref315,ref316}. GANs utilize CNNs as key components to model the generator and discriminator (See Fig. \ref{fig:GAN}) \cite{ref77}\cite{ref83}\cite{ref84}.
The generator is a CNN that generates new samples, such as realistic images, while the discriminator is another CNN that aims to distinguish between real and fake samples \cite{ref77}. These networks are trained adversarially, where the generator's goal is to produce samples that deceive the discriminator, and the discriminator's goal is to become better at distinguishing real from fake \cite{ref71}\cite{ref84}.

GANs with convolution have revolutionized the field of image generation and have produced impressive results in generating high-quality images and realistic textures \cite{ref266,ref267,ref268,ref269,ref270,ref271,ref272,ref273,ref274,ref275}. They have also been extended to other domains like NLP, audio generation, and video synthesis.
This technology has also been applied to other areas such as medical imaging, where GANs have been used to generate high-resolution and accurate images for diagnostic purposes. Additionally, GANs have shown promising results in the field of data augmentation, where they can generate synthetic data to increase the size and diversity of training datasets, improving the performance of machine learning models. 

For example, in the field of image generation, GANs with convolutional networks have been used to create realistic images of non-existent landscapes. The generator network creates visually convincing images, while the discriminator network learns to identify any flaws or inconsistencies in these generated images, pushing the generator to improve its output. This adversarial training process ultimately leads to the creation of high-quality and believable images that are indistinguishable from real photographs.

\subsection{Vision Transformers and Self-Attention Mechanisms}
Through the use of self-attention mechanisms \cite{ref85}, Vision Transformers \cite{ref244,ref245,ref246,ref247,ref248,ref249,ref250,ref251,ref252,ref253,ref254,ref255,ref256,ref257,ref258,ref259,ref260,ref261,ref262,ref263,ref264,ref265} represent an important evolutionary step away from traditional computer vision architectures \cite{ref86,ref87} . Rather than solely relying on convolutional filters to process visual inputs, as has predominantly been the case, they segment images into distinct finite parts known as patches \cite{ref87}. Each patch focuses on and extracts features from a different localized region of the photographic scene. This division of images into discrete patches is a major conceptual divergence from how most previous approaches operate.

In conclusion, advanced convolutional techniques have significantly expanded the capabilities of CNNs and revolutionized various fields like CV, image synthesis, and NLP. From transposed convolution for upsampling to capsule networks for handling spatial hierarchies, these innovations have enhanced the efficiency, robustness, and expressiveness of CNNs, making them powerful tools for a wide range of applications. Moreover, recent advancements, such as NAS and GANs, continue to drive progress in the field of DL and hold promise for further breakthroughs in the future. 

\section{Applications of Different Convolution Types}
\label{sec:Applications}
\begin{figure}[t]
    \centering
    \includegraphics[width=\linewidth]{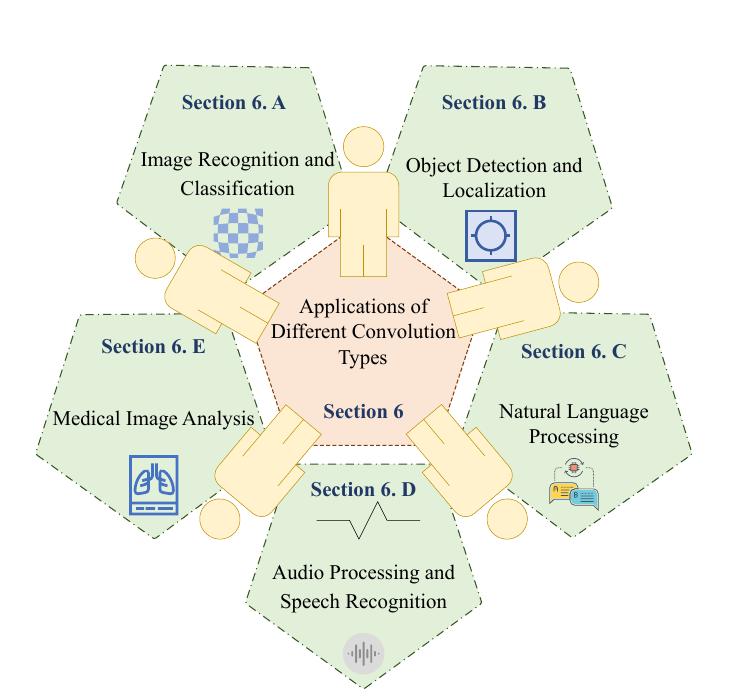}
    \caption{The  applications of CNN techniques which we have discussed in Section VI}
    \label{fig:Section 6}
\end{figure}

We provide a thorough overview of the numerous applications of different convolutional types in this section (See Fig. \ref{fig:Section 6}). Table \ref{tab:Main Applications} provides a brief but comprehensive overview of these applications.
Convolutions of various types are used in a variety of contexts, demonstrating the flexibility and strength of CNNs. Convolutional techniques enable machines to understand and interact with complex data, facilitating advancements in a variety of fields and enhancing our daily lives. Examples include image recognition, object detection, NLP, and medical image analysis.
\subsection{Image Recognition and Classification }
There are many uses for CNNs, including image recognition and classification. Traditional 2D convolutions are especially useful in these applications. They make it possible for deep learning models to accurately classify images into various groups and learn crucial features from images. The network's convolutional layers recognize edges, textures, and shapes. The pooling layers reduce the size of the image while preserving the data needed for classification. Image recognition and classification are used for various tasks, including optical character recognition (OCR) \cite{ref203,ref204,ref205,ref206,ref207,ref208,ref209,ref210,ref211}, classifying different animal species, and recognizing handwritten numbers \cite{ref88}. In competitions like ImageNet, CNNs have displayed impressive results, showcasing their abilities for handling wide image classification \cite{ref89}.

\begin{table*}[]
\centering
\setlength\extrarowheight{5pt}
\caption{The Compact Table Highlights the Main Applications of Each Convolution Type}
\label{tab:Main Applications}
\resizebox{\textwidth}{!}{%
\begin{tabular}{cccccc}
\hline
\rowcolor[HTML]{C0C0C0} 
\textbf{Convolution Type} &
  \textbf{Traditional 2D Convolutions} &
  \textbf{1D Convolutions} &
  \textbf{3D Convolutions} &
  \textbf{Dilated Convolutions} &
  \textbf{Grouped Convolutions} \\ \hline
\textbf{Image Recognition} &
  Image categorization &
  Time series analysis &
  Action recognition &
  Image segmentation &
  Real-time recognition \\ \hdashline
\textbf{Object Detection} &
  Object detection &
  Event detection &
  3D object detection &
  Semantic segmentation &
  Efficient detection \\ \hdashline
\textbf{NLP} &
  Sentiment analysis &
  Text classification &
  Textual entailment &
  \begin{tabular}[c]{@{}c@{}}Hierarchical \\  document classification\end{tabular}  &
  Parameter reduction \\ \hdashline
\textbf{ASPR} &
  \begin{tabular}[c]{@{}c@{}}Voice activity \\  detection\end{tabular}   &
  Speech recognition &
  \begin{tabular}[c]{@{}c@{}}Environmental sound \\  classification\end{tabular}   &
  \begin{tabular}[c]{@{}c@{}}Robust speech \\  recognition\end{tabular} &
  \begin{tabular}[c]{@{}c@{}}Low-latency \\  speech recognition\end{tabular}   \\  \hdashline
\textbf{Medical Image Analysis} &
  Tumor segmentation &
  ECG signal processing &
  \begin{tabular}[c]{@{}c@{}}Brain Tumor \\  Segmentation\end{tabular}  &
  \begin{tabular}[c]{@{}c@{}}Enhanced image \\  segmentation\end{tabular} &
  Faster medical analysis \\ \hline
\end{tabular}%
}
\end{table*}
\subsection{Object Detection and Localization}
Multiple objects within an image must be located and identified during object detection \cite{ref90}. In this application, both conventional 2D convolutions and 3D convolutions are crucial \cite{ref178,ref179,ref180,ref181,ref182,ref183,ref184,ref185,ref186,ref187,ref188,ref189,ref190,ref191}. While 3D convolutions are used for video object detection, 2D convolutions are used to process individual image frames. CNNs can detect objects at different scales and aspect ratios thanks to their region proposal mechanisms and anchor-based methods \cite{ref192,ref193,ref194,ref195,ref196,ref197,ref198,ref199,ref200,ref201,ref202}. 

Accurate localization of object bounding boxes is made possible by the use of pooling layers and convolutional sliding windows. Robotics, surveillance technology, and autonomous vehicles all use object detection to better understand and interact with their surroundings \cite{ref91}\cite{ref92}. 

\subsection{Natural Language Processing }
For sequential data, such as text processing and sentiment analysis, NLP uses 1D convolutions. 1D convolutions are used in NLP applications to extract pertinent patterns and relationships from sentences, enabling models to understand semantic meaning and context \cite{ref212,ref213,ref214,ref215,ref216}. Sentiment analysis for understanding customer opinions, named entity recognition to extract specific information from text, and text classification to classify news articles or product reviews are examples of NLP applications using 1D convolutions. Applications like machine translation and text summarization have benefited from the successful integration of CNNs and recurrent neural networks (RNNs). 

\subsection{Audio Processing and Speech Recognition}
Audio Processing and Speech Recognition (APSR) benefit from 1D convolutions, which analyze and process sequential audio data such as speech signals or audio waveforms \cite{ref217,ref218,ref219,ref220,ref221,ref222,ref223}. By extracting temporal patterns and acoustic features, CNNs can learn to recognize spoken words and transcribe audio into text. SR systems, often built upon convolutional and recurrent neural networks, enable voice assistants like Siri and Google Assistant to understand and respond to user commands. 

\subsection{Medical Image Analysis}
Medical image analysis involves the examination and interpretation of medical images, such as MRI scans, CT scans, and X-rays \cite{ref92}\cite{ref224,ref225,ref226,ref227,ref228,ref229,ref230,ref231,ref232,ref233,ref234,ref235,ref236,ref237,ref238,ref239,ref240,ref241}. In this domain, 3D convolutions and dilated convolutions are frequently used. 3D convolutions process volumetric medical data, allowing CNNs to extract spatial and contextual information for applications like tumor segmentation, organ localization, and disease classification \cite{ref92}\cite{ref93}. Dilated convolutions enhance feature extraction and semantic segmentation in medical images, enabling precise identification of abnormal tissues and structures. The applications of convolution types in medical image analysis have led to significant advancements in healthcare, assisting doctors in diagnosis and treatment planning. 

\section{Future Trends in CNN}
\label{sec:Future}
CNNs continue to be a hot topic of research and have achieved remarkable success in various CV applications. Future trends and open research questions in the field of CNNs are emerging as technology develops and deep learning techniques become increasingly complex.

The investigation of more effective architectures that can achieve comparable performance with fewer parameters and computational resources is one future trend in CNN research. How to make CNNs more interpretable is another unanswered research question, as the reasoning behind CNN decisions is frequently difficult to comprehend due to the internal complexity of these systems. Another crucial area for future research is finding ways to strengthen CNNs and make them less vulnerable to hostile attacks.

One active area of research looks at designing efficient CNN architectures optimized for edge and mobile computing. As CV moves from data centers to cameras, smartphones, and IoT at the network's edge, models need to operate within strict constraints on latency, memory, and power. Techniques including network pruning, compact operators, knowledge distillation, and adaptive quantization help derive lightweight CNN variants suitable for these low-resource scenarios \cite{ref121}. This focus on efficiency ties into work on improving CNN interpretability.

While today's complex CNNs achieve top accuracy, their decision-making remains poorly understood. Work on saliency mapping, activation clustering, modular CNNs, and other explanatory methods aims to shine light into the "black box" and address concerns around reliability, bias, and accountability - important considerations for safety-critical domains like healthcare. New types of CNN modules also aim to expand what these models can represent by incorporating flexible self-attention and capturing non-Euclidean structures.

A particularly compelling avenue involves tackling large-scale vision multimodal (LVM) challenges, which builds upon this work on expanding CNN capabilities. Vast datasets merging diverse visual media with language, audio, and other inputs present unprecedented complexity. However, they also offer unprecedented opportunities to develop general, comprehensive models of multisensory scene understanding.

\subsection{Interpretability and Explainability of CNNs}
The interpretability and explainability of CNNs is a significant open research question. Understanding the decision-making process of these models gets harder as CNNs get deeper and more complex. Particularly in critical applications like healthcare and autonomous systems, researchers are investigating ways to interpret and explain CNN predictions. To increase trust and reliability in CNN-based systems, methods such as attention visualization, saliency maps, and attribution methods seek to reveal which areas of the input contribute most to the model's conclusion.

\subsection{Incorporating Domain Knowledge}
Incorporating domain knowledge into CNN architectures is another important research direction. While CNNs have shown exceptional generalization abilities, they may not fully exploit domain-specific characteristics. Research focuses on developing architectures that can efficiently utilize domain knowledge or constraints, such as physics-based priors in medical imaging or geometric constraints in robotics, to improve performance and reduce data requirements.

\subsection{Robustness and Adversarial Defense}
Enhancing the robustness of CNNs against adversarial attacks remains a significant challenge. Adversarial attacks involve adding carefully crafted perturbations to inputs, leading to incorrect predictions by the CNN model. Researchers are investigating techniques for adversarial defense, such as adversarial training, robust optimization, and input transformations, to make CNNs more resilient against these attacks.

\subsection{Efficient Model Design}
When using CNNs on devices with limited resources, such as smartphones and edge devices, efficiency in terms of computation, memory, and power consumption is important \cite{ref242, ref243}. Creating lightweight architectures, knowledge distillation methods, and effective model compression techniques will be future trends in CNN research to decrease the model size and increase inference speed while maintaining accuracy.

\begin{table*}[]
\centering
\setlength\extrarowheight{3pt}
\caption{Comparison of Pruning Technique}
\label{tab:Pruning}
\resizebox{\textwidth}{!}{%
\begin{tabular}{cccccccc}
\hline
\rowcolor[HTML]{C0C0C0} 
\textbf{Technique} &
  \textbf{Sparsity Type} &
  \textbf{Pruning Granularity} &
  \textbf{Hardware Friendly} &
  \textbf{Accuracy Impact} &
  \textbf{Compression Ratio} &
  \textbf{Iterative Training} &
  \textbf{Requires Retraining} \\ \hline
Magnitude Pruning   & Unstructured & Weight level  & No  & Medium & 2-10x & Yes & No  \\
Filter Pruning      & Channel-wise & Filter level  & Yes & Low    & 5-10x & No  & Yes \\
Block Pruning       & Block-level  & Block level   & Yes & Low    & 2-5x  & No  & Yes \\
Network Slimming    & Channel-wise & Channel level & Yes & Low    & 2-5x  & Yes & Yes \\
Lottery Ticket      & Unstructured & Weight level  & No  & Low    & 2-10x & Yes & Yes \\
Iterative Magnitude & Unstructured & Weight level  & No  & Medium & 2-5x  & Yes & No  \\
Pruning-at-Init     & Channel-wise & Filter level  & Yes & Low    & 5-10x & No  & No  \\
One-Shot Pruning    & Channel-wise & Filter level  & Yes & Low    & 5-10x & No  & No  \\ \hline
\end{tabular}%
}
\end{table*}

Model compression techniques play a crucial role in designing efficient deep learning models suitable for deployment on resource-constrained edge devices. Several methods (See Table \ref{tab:Pruning}) have been proposed to reduce model size and computations without significantly impacting predictive performance. Network pruning and quantization are two widely used compression approaches\cite{ref102}\cite{ref103}.

Pruning techniques aim to sparsify neural networks by removing redundant connections with minimal impact on functionality \cite{ref121}. Early methods relied on unstructured pruning where connections were simply set to zero based on their magnitude or importance ranking. However, such arbitrary pruning leads to non-standard sparse matrices thereby preventing hardware acceleration. More recent structured pruning techniques induce channel-wise, filter-wise, or block-wise sparsity to yield compact models amendable to efficient implementations \cite{ref121,ref122, ref123, ref124}.

\begin{table*}[]
\centering
\setlength\extrarowheight{3pt}
\caption{Comparison of Quantization Technique}
\label{tab:quantization}
\resizebox{\textwidth}{!}{%
\begin{tabular}{cccccccc}
\hline
\rowcolor[HTML]{C0C0C0} 
\textbf{Technique} &
  \textbf{Quantization Level} &
  \textbf{Bit Width} &
  \textbf{Hardware Friendly} &
  \textbf{Accuracy Impact} &
  \textbf{Compression Ratio} &
  \textbf{Iterative Training} &
  \textbf{Requires Calibration} \\ \hline
Weight Quantization &
  Weight values &
  8-bit &
  Yes &
  Low &
  Up to 8x &
  No &
  Yes \\
\begin{tabular}[c]{@{}c@{}}Activation\\  Quantization\end{tabular} &
  Activations &
  8-bit &
  Yes &
  Low &
  Up to 8x &
  No &
  Yes \\
Tensor Quantization &
  Tensors &
  4-8 bit &
  Yes &
  Low &
  Up to 32x &
  No &
  Yes \\
Tensor Decomposition &
  Tensors &
  4-bit &
  Yes &
  Medium &
  Up to 32x &
  No &
  No \\
Huffman Coding &
  Weights &
  Variable &
  No &
  Low &
  Up to 10x &
  No &
  No \\
Log Quantization &
  Activations &
  1 bit &
  Yes &
  Low &
  Up to 16x &
  No &
  No \\
BNN Quantization &
  \begin{tabular}[c]{@{}c@{}}Weights/\\ Activations\end{tabular} &
  1 bit &
  Yes &
  High &
  Up to 32x &
  Yes &
  Yes \\
\begin{tabular}[c]{@{}c@{}}Floating Point\\  Quantization\end{tabular} &
  \begin{tabular}[c]{@{}c@{}}Weights/\\ Activations\end{tabular} &
  16-bit &
  Yes &
  Low &
  Up to 2x &
  No &
  No \\ \hline
\end{tabular}%
}
\end{table*}

Filter pruning refers to removing entire convolutional filters, thereby achieving channel-wise sparsity \cite{ref116}\cite{ref123}. It has been shown that up to 90\% of filters can be removed from VGG16 without accuracy degradation. One method, termed “Pruning-at-Initialization” prunes filters with the lowest sum values at the start of training itself. Alternatively, “One-Shot” prunes filters once based on their first-order Taylor expansion. These filter-level pruning methods lead to uniform sparsity across layers and reduce computation by \~5x.

Another structured approach is to prune blocks of connections rather than individual weights \cite{ref124}. For example, in “Block Level Pruning”, a number of convolution blocks are removed from blocks 1, 2, and 3 of ResNet50, reducing computations without retraining. The block structure ensures layout sparsity, maintaining original convolution block shapes for hardware friendliness. Network slimming is a channel-pruning method that enforces L1-norm regularization during training itself to gradually remove channels with low importance scores.

In unstructured variants, magnitude-based pruning removes weights below a threshold while iterative magnitude pruning alternates between weight updates and pruning based on a dynamic threshold \cite{ref121}\cite{ref125}. These maintain sparsity throughout the architecture but induce non-zero filler weights. Lottery ticket hypothesis experiments have demonstrated that dense, randomly-initialized, sub-networks can achieve the accuracy of their original networks if trained in isolation.

Apart from pruning, quantization is another effective technique to compress models (See Table \ref{tab:quantization}). Weight and activation quantization methods map weights/activations to a small set of discrete values, reducing the number of bits required for representation \cite{ref114}\cite{ref115}. For example, 8-bit quantization reduces model size by 4x without accuracy loss for many architectures. Tensor decomposition-based quantization further compresses models by decomposing weight tensors into low-rank approximations.

Some recent works have combined multiple compression approaches in a multi-stage pipeline. One example jointly employs weight quantization, pruning, and Huffman coding on ResNet50, achieving over 10x compression with a minor accuracy drop. Another uses a two-phase pipeline consisting of filtering-based pruning followed by quantization to design efficient MobileNet variants. Such composite methods achieve better accuracy-efficiency tradeoffs than individual techniques alone.

In conclusion, network pruning and quantization offer promising avenues to design compact models for edge and mobile applications. While early methods relied on unstructured sparsing, recent techniques induce structure for hardware friendliness. Looking ahead, continued research on model compression holds the key to facilitating the adoption of deep learning across myriad resource-constrained environments.

\subsection{Multi-Task Learning and Transfer Learning}
CNNs are well suited for multi-task learning, in which a single model is trained to carry out several related applications concurrently \cite{ref162,ref163,ref164,ref165,ref166,ref167,ref168,ref169,ref170,ref171,ref172,ref173,ref174,ref175,ref176,ref177}. The need for large amounts of labeled data for each individual task is being reduced as researchers investigate ways to take advantage of shared representations across applications and enhance generalization by transferring knowledge learned from one task to another \cite{ref147,ref148,ref149,ref150,ref151,ref152,ref153,ref154,ref155,ref156,ref157,ref158,ref159,ref160,ref161}.

\subsection{Integration with Uncertainty Estimation}
Understanding model uncertainty is essential for safety-critical applications. Integrating uncertainty estimation into CNNs would allow models to quantify their confidence in predictions and prevent costly errors, which is an area of open research. To improve the uncertainty measures in CNNs, researchers are investigating Bayesian neural networks (BNNs), dropout-based uncertainty estimation, and Bayesian optimization techniques.

\subsection{Generalization to Small Data Regimes}
A constant problem in the CNN research area is the generalization to small data regimes, where labeled training data are hard to come by. Essentially using data from related applications or domains, techniques like transfer learning, few-shot learning, and meta-learning work to increase CNNs' capacity to learn from sparse data.

\subsection{Evolution of Language Models and Multimodal LLMs}
In recent epochs, the domain of large language models (LLMs) for natural language processing has witnessed a precipitous progression. Prototypes such as BERT, GPT-3, and PaLM have demonstrated exceptional aptitude in language apprehension and generation, courtesy of self-supervised pretraining on voluminous text corpora \cite{ref85}. As LLMs expand in magnitude and range, incorporating additional modalities beyond text is a burgeoning field of study. Multimodal LLMs strive to amalgamate language, vision, and other sensory inputs within a singular model architecture. They hold the potential to attain a more holistic understanding of the world by concurrently learning representations across diverse data types \cite{ref96}. A significant hurdle is the effective fusion of the strengths of CNNs for computer vision and transformer architectures for language modeling.

One strategy involves employing a dual-stream architecture with distinct CNN and transformer encoders interacting via co-attentional transformer layers \cite{ref97}. The CNN extracts visual features from images, providing contextual information that can guide language generation and comprehension. The transformer architecture models the semantics and syntax of text. Their interaction enables the generation of captions based on image content or the retrieval of pertinent images for textual queries. Alternative methods directly incorporate CNNs within the transformer architecture as visual token encoders that operate with text token encoders \cite{ref98}. The CNN projections of image patches are appended to text token embeddings as inputs to the transformer layers. This unified architecture allows for end-to-end optimization of parameters for both vision and language tasks. Self-supervised pretraining continues to be vital for multimodal LLMs to learn effective joint representations before downstream task tuning. Contrastive learning objectives that predict associations between modalities have proven highly effective \cite{ref99}. Models pre-trained on large datasets of image-text pairs have demonstrated robust zero-shot transfer performance on multimodal tasks.

As multimodal LLMs increase in scale, the efficient combination of diverse convolution types and attention mechanisms will be crucial. Compact CNN architectures could help to reduce the cost of computing. Sparse attention and memory compression techniques can assist with scalability.


\section{Performance and Efficiency Consideration}
\label{sec:Performance}
Considerations for performance and efficiency (See Figs. 17-20) in CNNs are critical in developing high-performing and resource-efficient models. Researchers can make informed decisions about optimizing their CNN architectures for various applications and deployment scenarios by analyzing computational complexity, trade-offs between accuracy and speed, memory requirements, and benchmarking on standard datasets.

\subsection{Computational Complexity of Different Convolutions}
The computational complexity of different convolutional techniques (See Table \ref{tab:evaluation}) is a critical aspect to consider when designing CNNs. It refers to the amount of computation required to perform a convolution operation on input data. The computational complexity is influenced by various factors, including the size of the input data, the size of the convolutional filters, and the number of channels in the feature maps. 

Traditional convolutional layers, such as the standard convolution and depthwise separable convolution, generally have higher computational complexity compared to other techniques. This is because they involve a large number of convolution operations, especially when dealing with high-resolution images or complex data. On the other hand, techniques like pointwise convolution and transposed convolution tend to have lower computational complexity, making them more suitable for certain resource-constrained applications. 

Understanding the computational complexity of different convolution types is crucial for optimizing the performance of CNNs. By selecting convolution techniques that align with the available computational resources, researchers can build efficient models that achieve a good balance between accuracy and speed.

As illustrated in Figs. \ref{fig:lenet} to \ref{fig:resnet} the Adam optimizer performed well, as evidenced by key observations \textcircled{1} through \textcircled{6}, in both accuracy and loss metrics. Overall, the use of CNN techniques such as VGG, ResNet, and LeNet resulted in improved accuracy and reduced loss.

Also, as depicted in Figure \ref{fig:cpu}, and based on key observation \textcircled{1},\textcircled{2}, and \textcircled{3}, it is evident that the Adam optimizer exhibits less CPU usage in comparison to five other optimizers - RMSprop, Adamax, Adagrad, SGD, and Nadam. This observation holds true when using LeNet-5, VGG16, and ResNet-50. Additionally, the memory usage of the Adam optimizer is among the lowest (See key observation \textcircled{4}).

\begin{figure}[t]
    \centering
    \includegraphics[width=\linewidth]{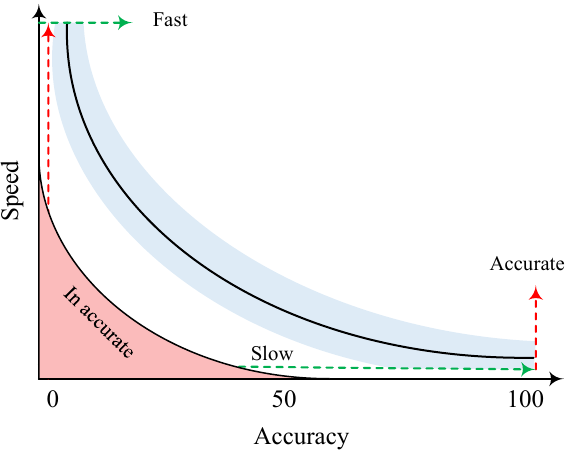}
    \caption{The trade-off curve between accuracy and speed of a deep learning model \cite{ref75}}
    \label{fig:tradeoff}
\end{figure}
\begin{table}[t]
\centering
\caption{Comparison on LeNet-5, VGG16, and ResNet-50 with 7 types of optimizers on Cifar-10 dataset, CU: CPU Utilization, MU: Memory Utilization}
\label{tab:evaluation}
\resizebox{\columnwidth}{!}{%
\begin{tabular}{cccccc}
\hline
\rowcolor[HTML]{C0C0C0} 
\textbf{Optimizer Type} & \textbf{CNN Model} & \textbf{Accuracy} & \textbf{Loss} & \textbf{CU} & \textbf{MU} \\ \hline
                                   & LeNet-5   & 0.547  & 1.277  & 71   & 50.7 \\
                                   & VGG16     & 0.87   & 0.776  & 57   & 55.7 \\
\multirow{-3}{*}{\textbf{SGD}}     & ResNet-50 & 0.789  & 1.1212 & 63   & 53.4 \\
                                   & LeNet-5   & 0.629  & 1.153  & 46.2 & 44.4 \\
                                   & VGG16     & 0.805  & 0.821  & 54.2 & 51.4 \\
\multirow{-3}{*}{\textbf{Adam}}    & ResNet-50 & 0.760  & 1.016  & 60.5 & 51.9 \\
                                   & LeNet-5   & 0.624  & 1.22   & 58.3 & 57.6 \\
                                   & VGG16     & 0.776  & 1.109  & 61.1 & 63.5 \\
\multirow{-3}{*}{\textbf{NAdam}}   & ResNet-50 & 0.789  & 0.89   & 66.4 & 57.8 \\
                                   & LeNet-5   & 0.605  & 1.288  & 50.3 & 42.9 \\
                                   & VGG16     & 0.755  & 22.286 & 61.2 & 49.7 \\
\multirow{-3}{*}{\textbf{RSMProp}} & ResNet-50 & 0.78   & 1.151  & 61.7 & 49.4 \\
                                   & LeNet-5   & 0.603  & 1.132  & 69.8 & 56.7 \\
                                   & VGG16     & 0.8506 & 0.885  & 55.8 & 64.2 \\
\multirow{-3}{*}{\textbf{Adamax}}  & ResNet-50 & 0.8123 & 1.002  & 62.1 & 56.1 \\
                                   & LeNet-5   & 0.412  & 1.65   & 67.6 & 44.4 \\
                                   & VGG16     & 0.822  & 0.708  & 55.3 & 50.3 \\
\multirow{-3}{*}{\textbf{AdaGrad}} & ResNet-50 & 0.75   & 0.999  & 62.4 & 50.6 \\ \hline
\end{tabular}%
}
\end{table}
\subsection{Trade-offs between Accuracy and Speed}

One of the key challenging aspects of designing CNNs is balancing model accuracy and inference speed (see Fig. 16). The inference time increases as the complexity of convolutional layers increases to capture more complex features. Using simpler convolutional techniques, on the other hand, may result in lower accuracy.  
The depth and width of the network, the number of parameters, the choice of convolutional techniques, and the hardware on which the model is deployed all have an impact on the trade-offs between accuracy and speed. For real-time applications or resource-constrained environments, sacrificing some accuracy to achieve faster inference may be necessary.
\begin{figure}[t]
    \centering
    \includegraphics[width=\linewidth]{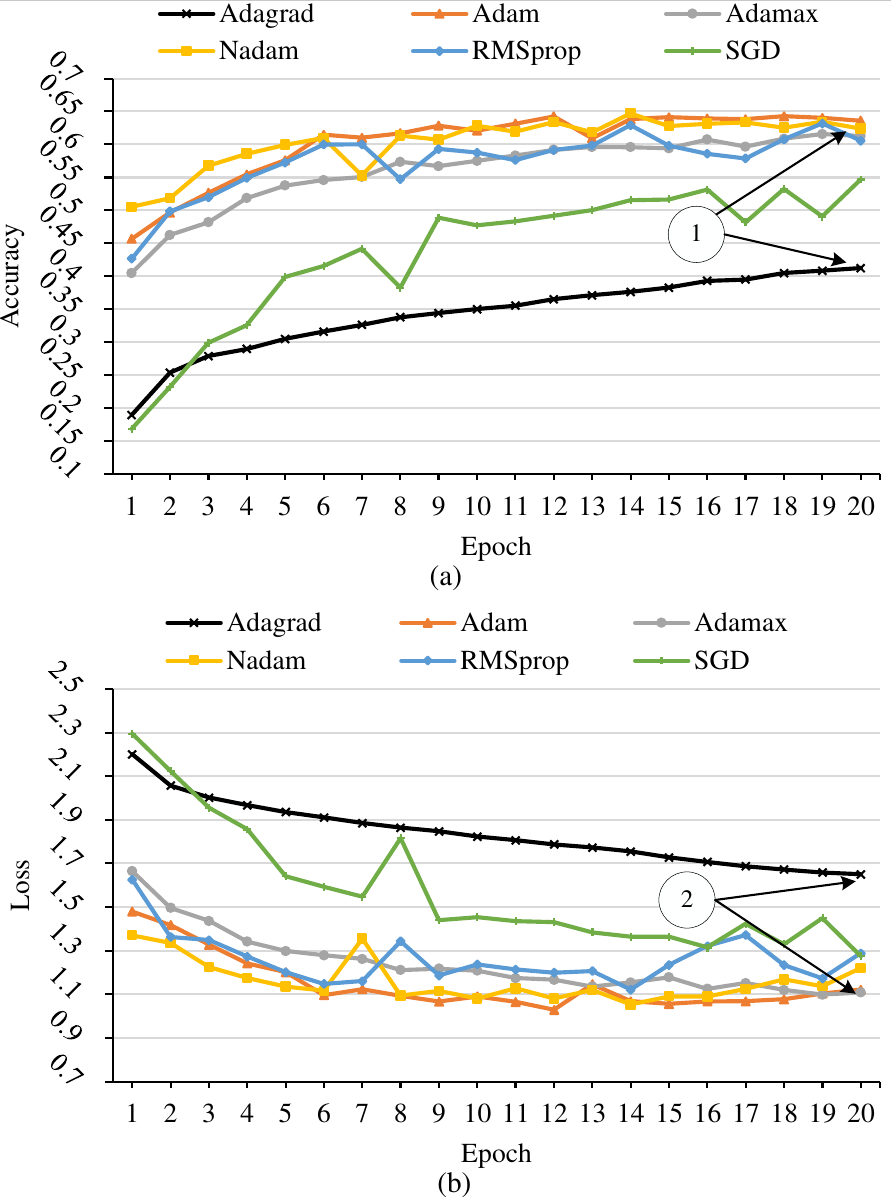}
    \caption{Comparison of various optimizers on LeNet-5 with Cifar-10 dataset. a) represents the accuracy of LeNet-5 architecture, b) represents loss of LeNet-5 architecture}
    \label{fig:lenet}
\end{figure}

Model pruning, quantization, and low-rank approximations are commonly used by researchers to reduce model size (See Section \ref{sec:Future} -> Subsection D) and improve inference speed without significantly compromising accuracy. Furthermore, attention-based convolutions and other techniques that prioritize important regions of the input can be used to focus computational efforts where they are most needed, improving the balance between accuracy and speed even further. 

\begin{figure}[t]
    \centering
    \includegraphics[width=\linewidth]{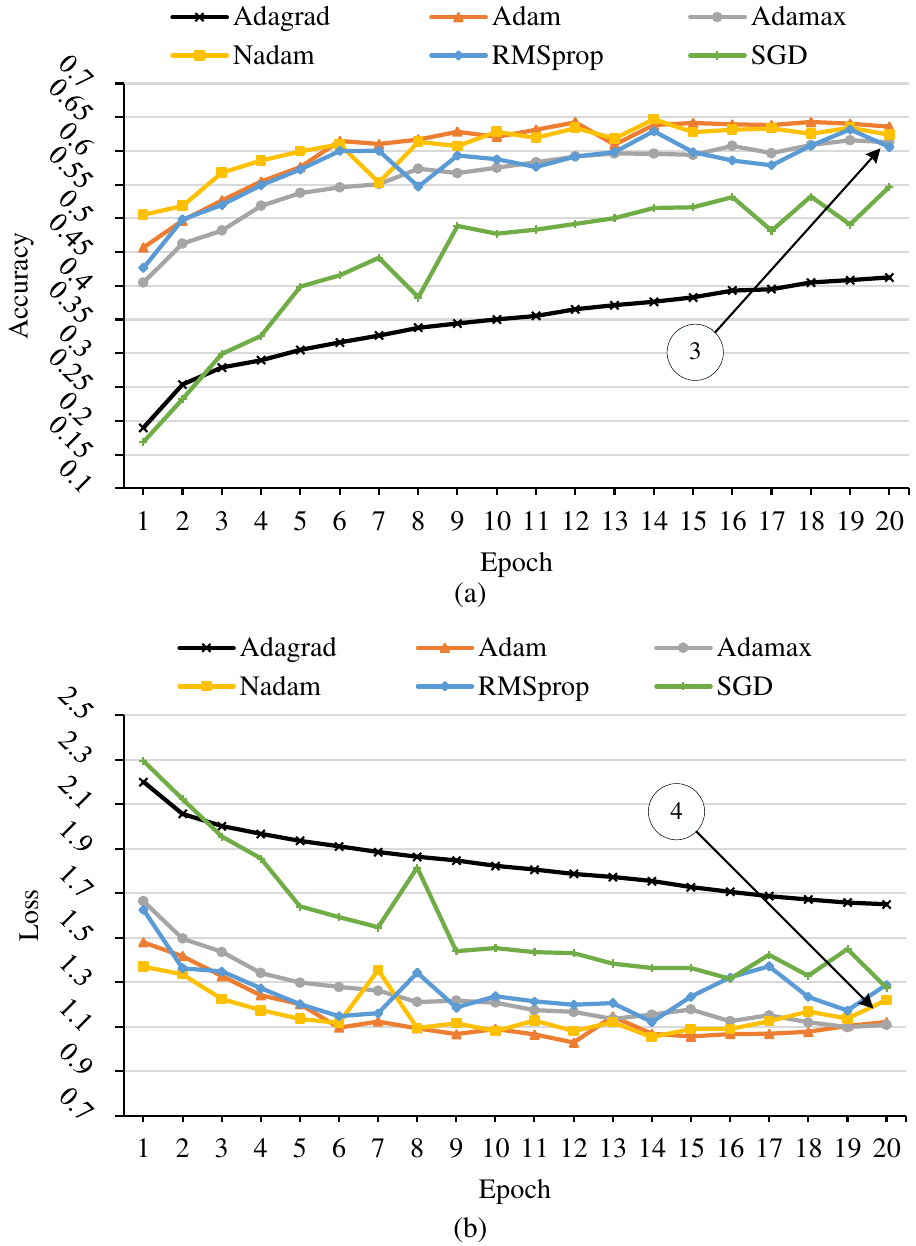}
    \caption{Comparison of various optimizers on VGG16 with Cifar-10 dataset. a) represents the accuracy of VGG16 architecture, b) represents loss of VGG16 architecture with various range of optimizers}
    \label{fig:vgg}
\end{figure}
\begin{figure}[t]
    \centering
    \includegraphics[width=\linewidth]{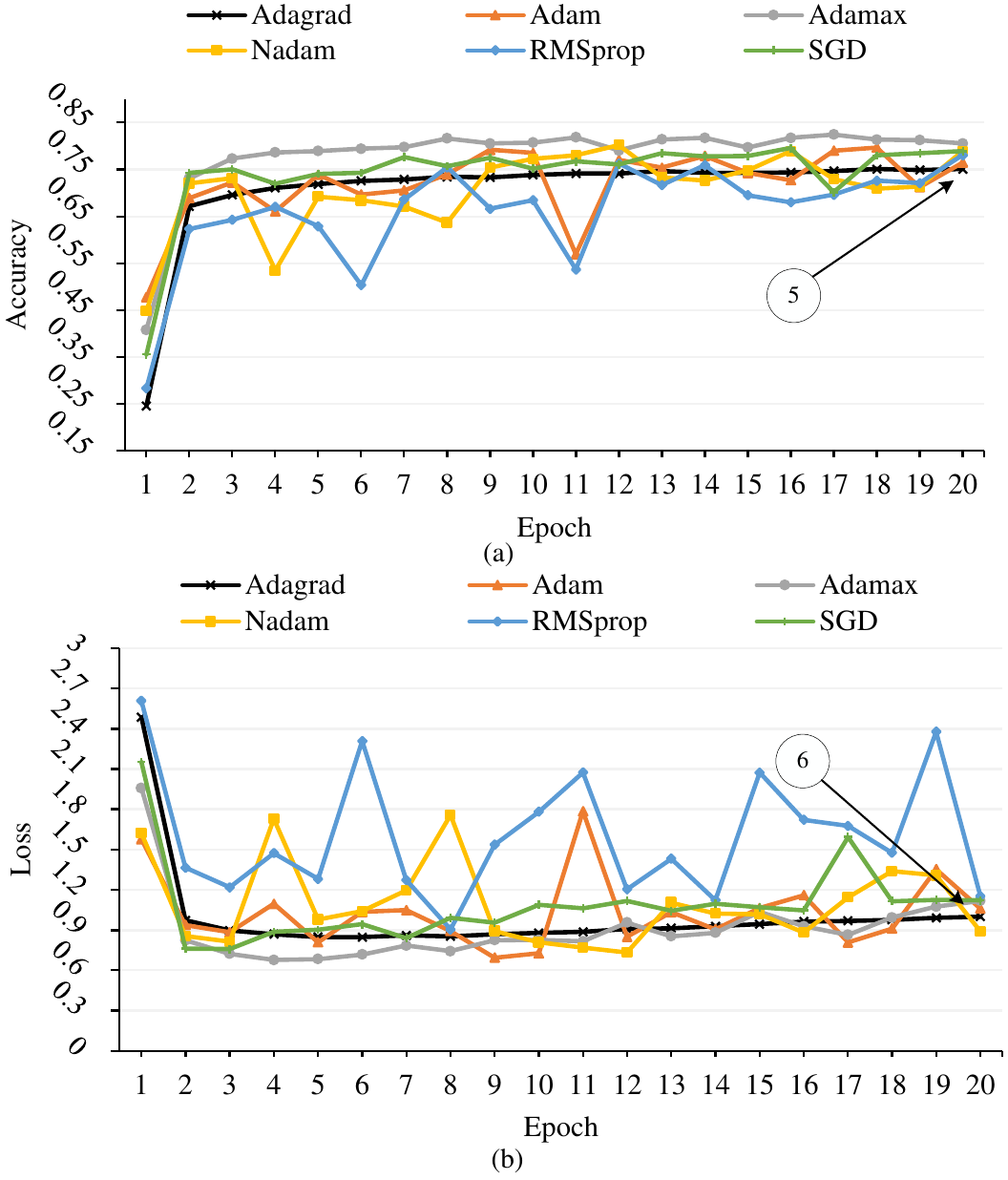}
    \caption{Comparison of various optimizers on ResNet-50 with Cifar-10 dataset. a) represents the accuracy of ResNet-50 architecture, b) represents loss of ResNet -50 architecture}
    \label{fig:resnet}
\end{figure}

\subsection{Memory and Storage Requirements}
\begin{figure}[t]
    \centering
    \includegraphics[width=\linewidth]{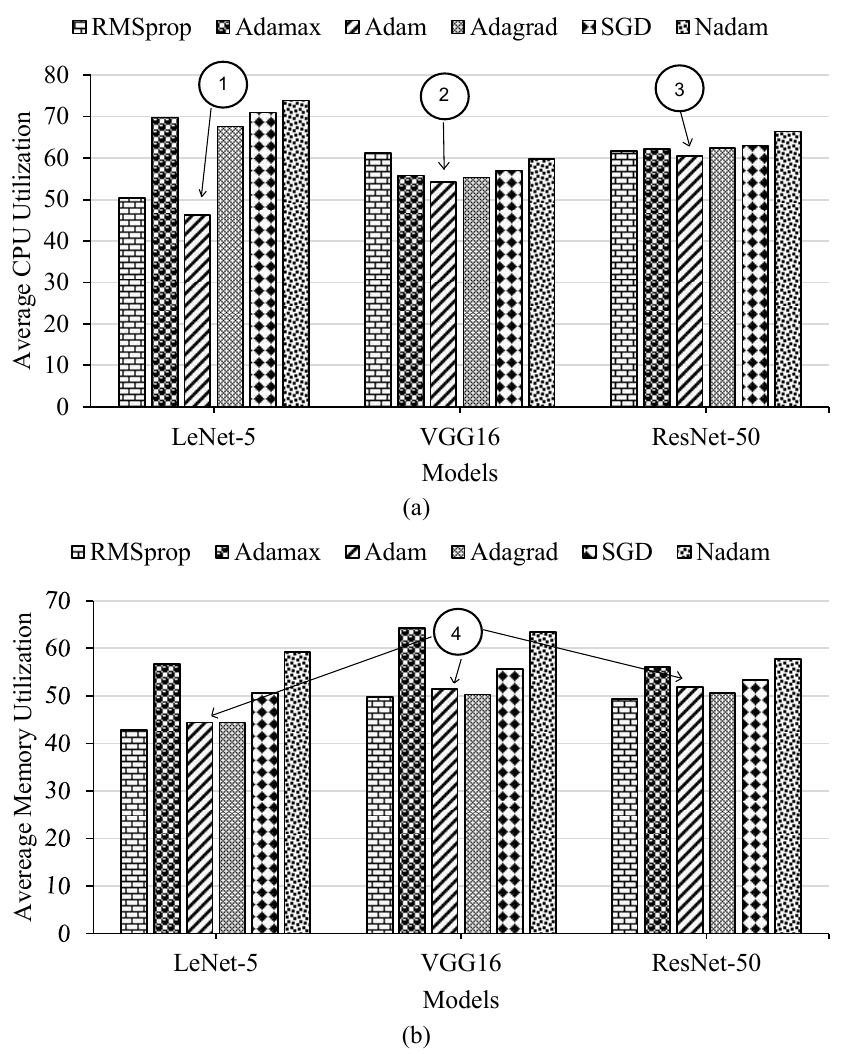}
    \caption{The CPU and Memory Utilization used by each model. a) The Average CPU Utilization of LeNet-5, VGG16, and ResNet-50 with six types of optimizer (Better value Recognition depends on use-case), b) The Average Memory Utilization of LeNet-5, VGG16, and ResNet-50 with six types of optimizer (Better value Recognition depends on Usecase)}
    \label{fig:cpu}
\end{figure}
Memory and storage requirements are crucial considerations in deep learning, especially when deploying models on edge devices or in cloud environments with limited resources. Convolutional models, particularly those with a large number of layers and parameters, can demand substantial memory and storage resources during training and inference. 

Traditional convolutional layers often have higher memory requirements due to the need to store intermediate feature maps and gradients during backpropagation. Depthwise separable convolutions and pointwise convolutions can reduce memory usage by reducing the number of parameters and intermediate feature maps. Memory-efficient CNN design involves strategies like using smaller batch sizes, employing mixed-precision training, and optimizing memory usage during inference. Additionally, model compression techniques, such as knowledge distillation and model quantization, can significantly reduce the size of the model without significant loss in performance. 

\subsection{Benchmarking on Standard Datasets}
Benchmarking convolutional techniques on standard datasets is a crucial step in evaluating their performance and efficiency. Standard datasets, such as ImageNet \cite{ref95} for image recognition or COCO \cite{ref94} for object detection, provide a common ground for fair comparison of different models and techniques. By benchmarking convolutional techniques, researchers can objectively assess their effectiveness in various applications and compare their performance with state-of-the-art models. The benchmarks consider metrics like accuracy, inference speed, memory usage, and energy efficiency, allowing for a comprehensive evaluation of the models. 

Benchmarking helps the DL community identify the strengths and weaknesses of different convolutional techniques, paving the way for improvements and advancements. It also aids practitioners in selecting the most suitable convolutional techniques for their specific use cases and desired trade-offs between performance and efficiency.

\section{Frameworks and Libraries}
\label{sec:Frameworks}
This section will provide an overview of some of the popular platforms (See Table \ref{tab:framework}) available for developing deep learning applications. We will compare the frameworks from aspects like their architecture, programming models, supported hardware, and key features. Choosing the right tool is crucial for deep learning success. That's why exploring framework capabilities is key for researchers and engineers

\begin{table*}[]
\centering
\setlength\extrarowheight{3pt}
\caption{Comparison of existing popular frameworks and libraries}
\label{tab:framework}
\resizebox{\textwidth}{!}{%
\begin{tabular}{ccccccccc}
\hline
\rowcolor[HTML]{C0C0C0} 
{\color[HTML]{000000} \textbf{Aspect}} &
  {\color[HTML]{000000} \textbf{Caffe}} &
  {\color[HTML]{000000} \textbf{TensorFlow}} &
  {\color[HTML]{000000} \textbf{Keras}} &
  {\color[HTML]{000000} \textbf{PyTorch}} &
  {\color[HTML]{000000} \textbf{OpenCV}} &
  {\color[HTML]{000000} \textbf{Deeplearning4j}} &
  {\color[HTML]{000000} \textbf{MXNet}} &
  {\color[HTML]{000000} \textbf{Chainer}} \\ \hline
Year released &
  2013 &
  2015 &
  2015 &
  2016 &
  1999 &
  2014 &
  2015 &
  2015 \\
\begin{tabular}[c]{@{}c@{}}Programming \\ language\end{tabular} &
  C++/Python &
  Python, C++ &
  Python &
  Python &
  C++, Python, Java &
  Java, Scala &
  \begin{tabular}[c]{@{}c@{}}Python, C++, R,\\ Scala, Perl, Julia\end{tabular} &
  Python \\
License &
  BSD 3-Clause &
  Apache 2.0 &
  MIT &
  BSD 3-Clause &
  BSD 3-Clause &
  Apache 2.0 &
  Apache 2.0 &
  MIT \\
Model definition &
  Layered &
  Graph-based &
  \begin{tabular}[c]{@{}c@{}}Sequential \&\\  functional\end{tabular} &
  \begin{tabular}[c]{@{}c@{}}Dynamic \\
  computations graphs\end{tabular} &
  N/A &
  \begin{tabular}[c]{@{}c@{}}Sequential, compute\\  graphs\end{tabular} &
  Symbolic &
  \begin{tabular}[c]{@{}c@{}}Imperative and \\ declarative\end{tabular} \\
Ease of use &
  Intermediate &
  Intermediate &
  High &
  High &
  Low &
  Intermediate &
  Intermediate &
  High \\
Speed &
  Fast &
  Fast &
  Intermediate &
  Fast &
  Very fast &
  Fast &
  Fast &
  Fast \\
\begin{tabular}[c]{@{}c@{}}Support for \\ computer vision\end{tabular} &
  Very good &
  Excellent &
  Good &
  Excellent &
  Excellent (library) &
  Good &
  Goo &
  Good \\
Focus &
  \begin{tabular}[c]{@{}c@{}}Research\\  prototyping\end{tabular} &
  \begin{tabular}[c]{@{}c@{}}Production \&\\  research\end{tabular} &
  \begin{tabular}[c]{@{}c@{}}User-friendly \\ research\end{tabular} &
  \begin{tabular}[c]{@{}c@{}}Research \\ prototyping\end{tabular}  &
  \begin{tabular}[c]{@{}c@{}}Traditional \\ algorithms\end{tabular} &
  \begin{tabular}[c]{@{}c@{}}Enterprise \\ production\end{tabular} &
  \begin{tabular}[c]{@{}c@{}}Distributed training\\  at scale\end{tabular} &
  \begin{tabular}[c]{@{}c@{}}Intuitive high-level\\  APIs for research\end{tabular} \\
Distributed training &
  No &
  Yes &
  No &
  No &
  No &
  Yes &
  Yes &
  No \\
\begin{tabular}[c]{@{}c@{}}Model \\ deployment\end{tabular} &
  No &
  Yes &
  Yes &
  Yes &
  No &
  Yes &
  Yes &
  Limited \\
\begin{tabular}[c]{@{}c@{}}Hardware \\ support\end{tabular} &
  CPU, GPU &
  CPU, GPU, TPU &
  CPU, GPU &
  CPU, GPU, TPU &
  CPU, GPU &
  CPU, GPU &
  \begin{tabular}[c]{@{}c@{}}CPU, GPU, \\ TensorFlow\end{tabular}  &
  CPU, GPU \\
\begin{tabular}[c]{@{}c@{}}Documentation \\ quality\end{tabular} &
  Good &
  Excellent &
  Good &
  Excellent &
  Excellent &
  Good &
  Good &
  Good \\
\begin{tabular}[c]{@{}c@{}}Community \\ support\end{tabular} &
  Limited &
  Very active &
  Very active &
  Very active &
  Very active &
  Active &
  Active &
  Active \\ \hline
\end{tabular}%
}
\end{table*}

Table \ref{tab:framework} provides a comparison of several popular frameworks and libraries used in deep learning. It evaluates key aspects such as the year of release, programming languages supported, license type, model definition approaches, ease of use, speed, and focus or strength of each framework.

\subsection{Caffe}
Caffe was one of the earliest and most influential deep learning frameworks developed specifically for CV tasks \cite{ref131}. Released in 2013 by the Berkeley Vision and Learning Center (BVLC), Caffe made training convolutional neural networks much faster and more accessible. It has an easy-to-use C++/Python interface and was designed for speed and modularity. Caffe adopted a layered structure that greatly simplified model definition and training. This helped drive wider adoption and enabled researchers to rapidly iterate on vision models. While development has slowed in recent years, Caffe laid important groundwork and is still used for CV research.

\subsection{TensorFlow}
TensorFlow is an end-to-end open-source machine learning platform developed by Google \cite{ref132}. While not strictly a CV library, it has become one of the most popular and full-featured frameworks for building and training complex deep learning models. TensorFlow has excellent support for CV including pre-trained models, image loading and preprocessing utilities, object detection APIs, and more. Its flexibility has led to it being used for a very wide range of applications from image classification to semantic segmentation. TensorFlow also works seamlessly across CPUs and GPUs and can be easily deployed to production.

\subsection{Keras}
Keras is a high-level deep learning API that runs on top of popular frameworks like TensorFlow and CNTK \cite{ref133}. Keras was developed with a focus on user-friendliness, modularity and extensibility. It provides excellent abstractions and tools for developing and evaluating deep learning models quickly. For CV, Keras ships with the ImageDataGenertator for real-time data augmentation as well as pre-defined models like VGG16. It also supports popular CV tasks like image segmentation, object detection, and feature extraction through convenient APIs. Keras' simplicity has made it very approachable for developers.

\subsection{PyTorch}
PyTorch is an open-source deep learning platform developed by Facebook's AI Research Lab (FAIR) \cite{ref134}. In recent years it has emerged as a leading alternative to TensorFlow especially for CV and NLP applications. PyTorch has a strong focus on dynamic neural networks and shares similarities to MATLAB and Numpy. This makes for an intuitive, Pythonic interface that is well-suited to CV prototyping and experimentation. PyTorch also supports GPU/TPU training along with production deployment. It has a growing ecosystem of 3rd party libraries and community support. Like Keras, PyTorch integrates tightly with common CV tasks and datasets.

\subsection{OpenCV}
OpenCV (Open Source Computer Vision Library) is a popular CV and machine learning software library \cite{ref135}. While not specifically designed for deep learning, OpenCV contains many traditional CV algorithms and an extensive collection of image processing functions. These include capabilities like image filtering, morphological operations, feature detection and extraction, object segmentation, and face and gesture recognition among others. OpenCV integrates with deep learning frameworks and is frequently used for simpler CV tasks or as a pre-processing step before feeding data into neural networks.

\subsection{MXNet}
MXNet is a flexible, efficient, and scalable deep learning framework \cite{ref136}. Similar to TensorFlow, it supports a wide variety of programming languages and hardware environments. MXNet excels at distributed training and supports training models containing billions of parameters across hundreds of GPUs. It also includes algorithms for CV like image recognition, object detection, and semantic segmentation. Overall, MXNet strikes a good balance between flexibility, performance, and ease of use making it suitable for large-scale CV problems.

\subsection{Chainer}
Chainer is an open-source deep learning framework created by preferred networks in Japan \cite{ref137}. It provides straightforward neural network abstraction similar to Keras with imperative and declarative model definitions. Chainer focuses on intuitive high-level APIs combined with low-level performance. It includes CV functionality like image loading, augmentation, pre-trained models, and model export. Chainer supports GPU and multi-GPU training and deployment. Overall it provides a performant and productive environment for CV development.

\subsection{Deeplearning4j}
Deeplearning4j (Dl4j) was launched in 2014 as an open-source deep learning library for Java and Scala on the JVM \cite{ref138}. It enables large-scale distributed training on GPUs and CPUs. For CV tasks, Deeplearning4j offers tools like image loading, pre-trained models, model import from Keras and ONNX, and the samediff for dynamic model construction. Deeplearning4j focuses on production-ready deployment with capabilities like model serving, online prediction, and on-device inference via Android or iOS apps.

Overall, these libraries and frameworks represent the forefront of open-source tools transforming CV through deep learning. Each offers different strengths and tradeoffs between flexibility, performance, ease of use, and supported features. As CV tasks continue advancing, we can expect these projects to further incorporate state-of-the-art research while also lowering the barrier to development through improved tools and abstractions. CV is sure to remain a major application domain for deep learning innovation in both research and industry. 

\section{Main Research Fields}
\label{sec:Research}
\subsection{Image Classification}
Image classification was one of the earliest successes of CNNs. The seminal AlexNet achieved record-breaking results on the ImageNet challenge in 2012 by drastically improving upon prior techniques. Today, state-of-the-art CNNs for image classification routinely achieve human-level or better accuracy on standardized datasets. Architectures like ResNet, Inception, Xception, and EfficientNets optimize parameters, layer connectivity, and computation to classify thousands of object categories at superhuman performance levels \cite{ref52,ref53,ref56,ref276}. Beyond static images, video classification CNNs also extract spatial-temporal features to recognize complex activities and events.

\subsection{Object Detection}
Object detection is another major CV application that relies heavily on convolutional modeling. Two-stage detectors like Faster R-CNN and one-stage detectors like YOLO leverage region proposal networks and anchor boxes trained via priors to simultaneously localize and classify objects within images \cite{ref317,ref318,ref319,ref320,ref321,ref322,ref323,ref324,ref325,ref326,ref327,ref328,ref329,ref330,ref331}. Recent works further optimize speed and accuracy, enabling real-time object detection on billions of parameter models. Techniques like mobile object detection address embedded constraints by designing lightweight CNN backbones and feature extractors optimized for on-device inference \cite{ref332}.

\subsection{Image Segmentation}
Semantic segmentation tasks require dense pixel-level labeling of image content. FCN and U-Net CNNs employ skip connections and encoder-decoder mirrors to preserve spatial information across resolutions \cite{ref333,ref334,ref335,ref336,ref337,ref338,ref339,ref340,ref341,ref342,ref343,ref344,ref345,ref346,ref347,ref348}. PSPNet and DeepLab introduce pyramid spatial pooling modules to capture multi-scale contextual cues \cite{ref349}. GANs and conditional random fields further refine coarse segmentations from CNNs. Advances in medical imaging also apply segmentation CNNs to understand organ structures, localize pathologies, and aid diagnosis.

\subsection{Vision transformers}
Vision transformers have also emerged as a compelling alternative to traditional CNNs for CV tasks. Inspired by the success of language models, vision transformers divide images into discrete patches which are embedded and processed with self-attention. This allows them to capture long-range dependencies and multi-scale contextual information more effectively than CNNs. Models like ViT, DeiT, and Visual BERT demonstrate state-of-the-art results in tasks like image classification when pre-trained on large datasets \cite{ref350,ref351,ref352,ref353,ref354,ref355,ref356,ref357}. Research now focuses on optimizing transformer efficiency for real-time CV applications.

\subsection{One-shot/few-shot/Zero-shot learning}
One-shot and few-shot learning aim to address challenges posed by limited labeled training examples. Through metric learning and prototypical networks that learn robust representations from extensive base classes, models can effectively recognize new concepts from just one or a handful of examples without catastrophic forgetting \cite{ref358,ref359,ref360,ref361,ref362,ref363,ref364,ref365,ref366,ref367,ref368,ref369,ref370,ref371,ref372}. This opens up CV to new long-tailed and incremental learning paradigms. Matching networks and prototypical networks efficiently compare test samples to prototype representations of base classes to generalize from limited exposures.

Zero-shot learning emerges as a promising area where CNNs imagine possibilities beyond the limitations of labeled data \cite{ref373,ref374,ref375,ref376,ref377}. Descriptors like attributes or semantic relationships introduce inductive biases facilitating generalization without example. SAE, DeViSE, and contemporary models transfer knowledge by aligning embeddings between seen and unseen categories connected through auxiliary descriptors. Knowledge graphs also provide structural inductive biases through entity and relation modeling.

\subsection{Weakly-supervised learning}
Weakly supervised learning techniques also help alleviate dependence on labor-intensive annotations \cite{ref378,ref379,ref380,ref381,ref382,ref383}. Models can be trained end-to-end from weaker input signals like image-level tags or bounding box object locations instead of explicit pixel-level segmentation maps. Multi-instance learning approaches cluster image regions corresponding to each label to iteratively refine local predictions. Expectation-maximization (EM) and multiple instance learning jointly infer labels and recognize discriminative regions, enabling training from cheaper forms of weak supervision.

\subsection{Self-supervised/unsupervised learning}
Self-supervised learning has also gained vast attention in CV by enabling pre-training from sheer ubiquity of unlabeled visual data \cite{ref384,ref385,ref386,ref387,ref388,ref389,ref390,ref391,ref392,ref393}. Pretext tasks like predicting image rotations, solving jigsaw puzzles, or counting pixel colors allow models to learn rich visual representations applicable to downstream tasks. Recent contrastive self-supervised models like SimCLR, SwAV, and MoCo demonstrate that unlabeled pre-training rivals or exceeds supervised pre-training in various vision benchmarks, enabling more data-efficient fine-tuning or transfer to new problems.

\subsection{Lifelong/Continual learning}
Lifelong and continual learning aim to simulate open-world scenarios where models learn lifelong with non-stationary data distributions \cite{ref51}. Models must avoid catastrophic forgetting when presented with new classes or shifts in existing class definitions without revisiting historical data \cite{ref394,ref395,ref396,ref397,ref398,ref399,ref400,ref401,ref402,ref403}. Elastic weight consolidation and incremental moment matching regularization preserve knowledge while accommodating new tasks. Research now explores task-aware architectures, dual-memory systems, and replay buffers that emulate memory reconsolidation to model lifelong visual learning.

\subsection{Vision language model}
Vision-language models (VLMs) have also emerged at the intersection of NLP and CV by grounding language in visual contexts. Models fuse multimodal inputs through attention and generate captions conditioned on images, or localize and describe visual entities based on linguistic context. Large pre-trained models such as CLIP, ALIGN, and Oscar demonstrate exciting capabilities like zero-shot classification, question-answering (QA), and visual dialog with potential applications in education, assistive technologies, and more.

\subsection{Medical image analysis}
Medical imaging epitomizes the necessity of collaboration between deep learning and domain experts. Segmenting organs in volumetric scans, localizing anomalies across imaging modalities, and tracking patients longitudinally all leverage 3D/2D CNNs \cite{ref404,ref405,ref406,ref407,ref408,ref409,ref410,ref411,ref412,ref413,ref414,ref415,ref416,ref417,ref418}. Advanced models exploit anatomical priors by enforcing smoothness, and preservation of edges and surfaces in predictions. Self-supervision further enables pre-training from non-private data before fine-tuning target tasks. Model interpretation especially matters here to ensure trust among clinicians \cite{ref410,ref411,ref412,ref413}. Beyond diagnosis, CNNs can also simulate novel views to aid surgical planning. Efficiency additionally matters for on-device deployment and assisting underserved populations lacking infrastructure.

\subsection{Video understanding}

Beyond images, video understanding presents unique challenges in modeling spatial-temporal relationships across consecutive frames. C3D and I3D CNNs introduce 3D convolutions directly learning from video volumes. Advanced techniques in video captioning and action recognition fuse language models and attention to jointly reason about visual content and linguistic semantics over time. Self-supervised learning from large unlabeled video repositories also emerges as a promising pretraining paradigm before fine-tuning downstream tasks.

\subsection{Multi-task learning}
Multi-task learning aims to improve generalization by jointly training CNNs on multiple related tasks using shared representations. This has proven successful across numerous applications by leveraging commonalities while mitigating overfitting individual tasks' limited data \cite{ref419,ref420,ref421,ref422}. For example, YOLO trains object detection alongside other auxiliary predictions like segmentation and counting.

Multi-task CNNs outperform independent models in low-data regimes (See Section \ref{sec:Future} -> Sub-section G.) by borrowing statistical strength across related problems. Dense captioning localizes objects and describes scenes simultaneously. A single network predicts keypoints, normals, and semantic part segmentation. Deeper tasks benefit substantially from representations learned for more general shallow tasks.

Progressively growing into new problem spaces via related auxiliary objectives also prevents catastrophic forgetting. Self-supervised pre-training establishes features broadly useful across downstream tasks, including those without annotations. Measuring and maximizing modularity in multi-task architectures additionally reduces interference between domains.

Techniques like multi-granularity, multi-level, and heterogeneous multi-task learning further craft diverse objectives to progressively refine semantics captured at differing levels of granularity \cite{ref423,ref424,ref425,ref426}. Task relations range from independent, and cooperative where tasks improve each other, to completely shared exploiting identical representations. Properly designed, multi-task CNNs deliver state-of-the-art performance while improving generalizability, efficiency, and real-world applicability.

Multi-task models combine CNNs with other modalities like language. For captioning, CNN-RNN fusion grounds generated text within visual contexts. For retrieval, ranking loss trains CNN-LSTM encoders to map semantically aligned vision-text pairs to nearby embeddings. Multi-modal pre-training on enormous unlabeled multimedia collections has proven highly beneficial via self-supervised alignment of domains.

\subsection{6D vision}
6D vision aims to recover the full 6D pose (3D position, 3D orientation) of objects directly from monocular RGB images. This is a challenging problem due to the loss of depth information when projecting 3D scenes onto 2D images \cite{ref427,ref428,ref429,ref430,ref431,ref432}. Early works relied on CAD models and rendered synthetic data which lacked photorealism, while more recent approaches leverage large amounts of real training data.

CNN-based regression networks are commonly used which take images as input and directly predict the 6D pose values. PoseCNN showed this can achieve competitive accuracy to model-based regression if trained end-to-end on real data. Due to the complex, multi-modal nature of the target distribution, losses that ensure consistent predictions under different poses like reprojection or angular are beneficial.

Iterative refinement approaches first detect the object, then iteratively update the pose estimate based on 2D-3D correspondences. DeepIM predicts shape coefficients and refines using PnP. DPOD leverages deep features combined with geometric constraints in a RANSAC framework. Dense representations also help by reasoning about object parts independently.

Multi-view and RGB-D sensors provide additional cues to leverage. MVD helps constrain the problem by training separate networks for each view and fusing results. Using both RGB and depth as input allows Depth-PoseNet to lift 2D predictions to 3D space. Multitask models predicting bounding boxes, keypoints, and poses jointly demonstrate accuracies approaching marker-based motion capture.

\subsection{Neural Architecture Search}
Neural architecture search (NAS) aims to automate the design of neural networks leveraging the power of evolution and reinforcement learning. Rather than relying on human experts to laboriously craft CNN architectures, NAS approaches evolve architectures directly on target datasets and tasks. This has led to state-of-the-art vision models developed without human design choices \cite{ref433,ref434,ref435,ref436,ref437,ref438,ref439,ref440}.

Early NAS works explored various search spaces defined by units, operations, and connections between them. Combining concepts like pruning, sharing weights across child models during evolution helped scaling search to larger spaces \cite{ref295,ref299}. Performance predictors further reduced costs by guiding search towards promising regions. Novel methods evolved filters, activation functions, and batch normalization layers for particular domains.

Recent efforts evolve entire sections or blocks, expanding applicable search spaces. Single-path one-shot approaches drastically sped up search without compromising quality. ProxylessNAS found efficient mobile architectures directly on target devices. NAS approaches also discover non-CNN models suiting problems beyond CV.

Once identified, the best architectures can be trained from scratch to further improve upon proxy accuracies predicted during the search. Late phase evolution also enhances architectures initially identified, while architecture parameters themselves may evolve. Overall, NAS technologies continuously push forward state-of-the-art for vision tasks given diverse data, constraints, or objectives.

\subsection{Neural Architecture Transformer}
Neural architecture transformers (NAT) replace CNNs' fixed topology with self-attention, replacing convolutional filtering with axial self-attention \cite{ref436, ref441}. This increased flexibility allows modeling long-range pixel dependencies crucial for vision tasks like segmentation. Vil-BERT introduced a multi-stage training procedure enabling pre-trained models to learn visual representations as well as natural language tasks.

Early works divided input images into uniform patches processed independently by attention layers. More sophisticated designs aim to capture visual locality through hierarchical patch divisions better. Rotary positional embeddings and attention patterns help encode translation equivariance. Architectures like CoAtNet cascade blocks with increased resolution, improving accuracy and interpretability.

Multi-scale vision transformers (MViT) incorporate prior convolutional inductive biases in hybrid models jointly benefiting from attention and translation equivariance. Combining vision transformers with convolutional networks particularly benefits medical image segmentation leveraging anatomical priors. Swin Transformers introduces a shifted window mechanism to focus computation locally across higher-resolution feature maps.

Though still an emerging direction, neural architecture transformers open new pathways for CV by bringing the full generality of self-attention to bear on visual problems. Their continued development will surely impact future CV research by unlocking novel representational abilities. Alongside NAS, they hold promise for pushing boundaries through data-driven discovery operating directly within much broader algorithmic search spaces.

\subsection{Generative Models}
Generative models have made large strides in the area of CV through techniques like GANs and diffusion models \cite{ref442,ref443,ref444}. GANs pair a generator network against a discriminator network in an adversarial training procedure. This drives the generator to synthesize increasingly realistic fake images that can fool the discriminator.

GANs have produced impressive results generating photos that are near-indistinguishable from real images. Applications include image-to-image translation, super-resolution, and manipulating image attributes like style \cite{ref444,ref445,ref446,ref447}. However, GAN training remains tricky to stabilize. Issues like mode collapse require careful architecture and hyperparameter choices.

Diffusion models provide an alternative generative framework gaining popularity. They utilize denoising diffusion probabilistic models (DDPMs) which gradually corrupt data with Gaussian noise before reversing the process \cite{ref442,ref443,ref444,ref447,ref448,ref449}. During generation, the model adds noise to a blank canvas and then predicts the noise-reduced output iteratively. This diffusion process proves more stable than adversarial training.

Sampling from DDPMs follows an ancestral sampling approach regressing the noise at each step conditioned on the previous denoised output. Advanced techniques like score-based sampling further improve sample quality by maximizing the model's density rather than following ancestral noise. Generative diffusion models (GDMs) also maximize a denoising score objective specifically for a generation \cite{ref449}.

Diffusion models have proven highly effective at synthesizing crisp, detailed images across varied datasets. Large-scale vision diffusion models (LVMs) like DALL-E 2 and DALL-E 3 demonstrate unparalleled capabilities of generating images from text prompts, and can even fuse language and vision to answer trivia questions about synthetic images.

By generating synthetic training data, generative models also benefit downstream classification, detection, and segmentation tasks through data augmentation. As generative diffusion models continue advancing, they will surely establish new frontiers in CV domains ranging from image editing to scientific discovery through computational experimentation.

\subsection{Meta Learning}
Meta-learning, also known as learning to learn, aims to develop models that can rapidly adapt to new tasks and environments using only a few training examples. This is achieved by learning inductive biases about learning itself on a variety of related tasks during a meta-training phase. These biases are then leveraged during meta-test time on novel tasks \cite{ref450, ref451}.

In CV, meta-learning enables CNNs to generalize beyond the restrictions of limited labeled examples through fast adaptation. Model-agnostic meta-learning (MAML) trains initial model parameters such that a few gradient steps fine-tune into new tasks. This learns efficient parameter initialization rather than solutions for any specific task \cite{ref450,ref451,ref452,ref453,ref454,ref455,ref456,ref457}.

Metric-based approaches represent classes using prototypes that summarize inter/intra-class relationships independent of tasks \cite{ref450,ref451, ref452}. Matching networks compare new examples to prototypes, providing fast adaptation through learned metric space similarities. Meta-Dataset consolidates many few-shot image classification datasets, advancing state-of-the-art and evaluation protocols in this challenging zero/few-shot regime\cite{ref450,ref451, ref452, ref457}.

Self-supervised auxiliary tasks like prediction, rotation, and context modeling further enhance generalization when used alongside supervised meta-learning objectives. Temporal ensemble models aggregate diverse predictions over time from a generator network, improving robustness to noise and outliers. Reinforcement meta-learning successfully trains visuomotor policies for robotic control from only a handful of demonstrations.

\subsection{Federated Learning}
Federated learning (FL) enables distributed training across decentralized edge devices without exchanging private user data like images, videos or medical scans \cite{}. It aims to collaboratively learn a shared global model tailored to non-IID user distributions through coordinated local updates. This paradigm attracts increased interest due to growing concerns around data privacy and security.

FL trains a centralized CNN model through an iterative process where devices download the latest parameters, contribute updates computed over shards of local data, and then push weights back. A parameter server aggregates updates to globally improve the model. A key challenge arises from heterogeneity in non-IID data distributions, devices, and unreliable network connectivity. FedVision applies FL to object detection directly over fragmented client videos.

Techniques like personalized, multi-task, and meta-learning help address statistical heterogeneity in FL. Continual learning aspects prevent catastrophic forgetting when populations change over disseminated rounds. Differentially private algorithms and secure aggregation schemes ensure strong privacy in collaborative updates, advancing FL under stringent privacy constraints beyond vision to sensitive domains like healthcare.


\section{Discussion}
\label{sec:Discussion} 
We have methodically explored the various CNN variations that have become more and more popular in recent years across a wide range of application sectors through this thorough survey. Our goal in this discussion part is to summarize the most significant findings from our evaluation of the literature and offer an analytical viewpoint on significant problems regarding the development and prospects of this area of study.

Convolutional layers are well-suited for grid-like data types, like images because they have proven highly capable of capturing spatial relationships and extracting hierarchical patterns. At the core of CNNs, commonly used for computer vision tasks such as object identification and image classification, remain traditional 2D convolutions. However, as the field has evolved, additional specialized convolution approaches have emerged to handle different data modalities more effectively. One notable application of 1D convolutions is in sequential data domains including time series analysis and natural language processing. Their ability to capture temporal dependencies has enabled state-of-the-art accuracy on various language and audio processing problems. Likewise, 3D convolutions allow CNNs to effectively model volumetric medical images and video inputs by accounting for both spatial and temporal dimensions.

While basic convolution varieties such as 2D and 3D continue powering many top models, more efficient variants have also been developed. Dilated convolutions utilize dilations to widen receptive fields without loss of resolution, aiding high-level semantic tasks such as segmentation. Grouped convolutions offer a means of factorizing convolutions to dramatically reduce computation and memory usage, enabling large, deep architectures. However, their representational abilities may remain limited compared to standard convolutions for advanced analysis. Depthwise separable convolutions, as used in MobileNets, have achieved tremendous success in deploying efficient CNNs on embedded and mobile devices via their channel-wise decomposition.

In addition to novel convolution designs, the field is witnessing increasingly innovative integration of concepts from parallel research areas. For example, vision transformer models incorporate attention mechanisms to replace convolutional building blocks entirely, achieving strong results, especially on large datasets. Techniques like capsule networks aim to overcome CNN limitations through dynamic routing between feature vectors. Generative models such as Pix2Pix employ convolutional decoders to generate high-fidelity images from semantic maps or sketches. Advances in self-supervised learning provide alternative pretraining paradigms bypassing the need for vast annotated datasets.

Further combining of deep learning techniques seems poised to yield fruitful synergies. For instance, incorporating attention into convolutional pipelines could endow them with the benefits of both approaches. Moreover, self-supervised mechanisms may help the unsupervised discovery of interpretable convolutional filters well-suited to specific domains. Despite remarkable achievements, open challenges remain regarding robustness, sparse data scenarios, model interpretability, and trustworthiness. Future progress relies on close collaboration between academia and industry to define real-world needs and expand deep learning's positive societal impact.

Some convolution types have proven more enduring than others based on their flexibility and ability to adaptively fit diverse applications. While LeNet certainly played an instrumental pioneering role, more recent architectures better capture inherent data properties through principled network designs and optimizations. Meanwhile, innovation continues on all fronts, suggesting no single solution has emerged as definitive. Success hinges on judiciously combining innovations tailored to particular contexts rather than wholesale replacement of existing paradigms. 

A promising outlook envisions continued refinement of core CNN building blocks and their harmonious integration with new algorithmic concepts from self-supervised learning, attention mechanisms, and generative models. In conclusion, this survey highlights both the remarkable advances of convolutional neural networks to date and their vast unrealized potential through the future intersection of ideas across deep learning’s constantly evolving landscape.

\section{Conclusion}
\label{sec:Conclusion}
In this comprehensive study of different convolution types in deep learning, we have gained valuable insights into these techniques' diverse applications and strengths. CNNs have proven to be highly effective in various domains, ranging from image recognition to natural language processing. We compared various types of CNNS in various aspects, allowing us to understand their unique characteristics and advantages for specific tasks. Overall, this study emphasizes the importance of convolution in deep learning and its potential for future advances and improvements in artificial intelligence. Furthermore, the findings suggest that CNNs' versatility makes them suitable for various applications beyond traditional computer vision tasks. Furthermore, the study emphasizes the importance of additional research and development to optimize and refine these techniques for specific domains and tasks.

\section*{Acknowledgements}
This work was partially supported by the NYUAD Center for Artificial Intelligence and Robotics (CAIR), funded by Tamkeen under the NYUAD Research Institute Award CG010. This work was also partially supported by the project “eDLAuto: An Automated Framework for Energy-Efficient Embedded Deep Learning in Autonomous Systems”, funded by the NYUAD Research Enhancement Fund (REF).

\vspace{11pt}

\begin{IEEEbiography}[{\includegraphics[width=1in,height=1.25in, clip,keepaspectratio]{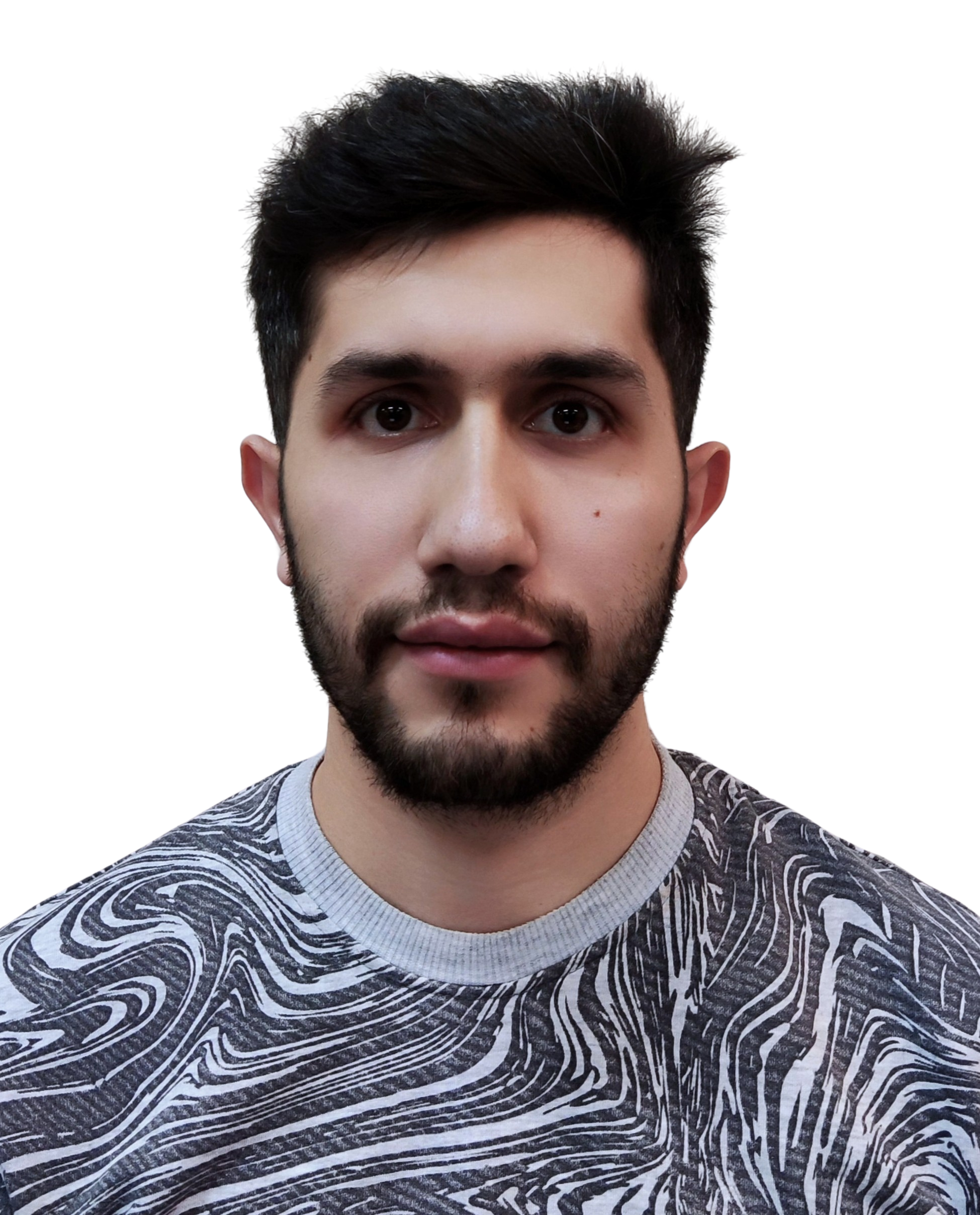}}]{Abolfazl Younesi} received a B.Sc. degree in computer engineering from the Tabriz University, Tabriz, Iran, in 2021. He is currently working toward an M.Sc. degree in computer engineering at Sharif University of Technology (SUT), Tehran, Iran, from Oct. 2021 until now. He is currently a member of the Embedded Systems Research Laboratory (ESR-LAB) at the Department of Computer Engineering, Sharif University of Technology. His research interests include the Internet of Things (IoT) and Cyber-Physical Systems (CPS), low-power design, machine learning, and computer vision.
\end{IEEEbiography}

\begin{IEEEbiography}[{\includegraphics[width=1in,height=1.25in, clip,keepaspectratio]{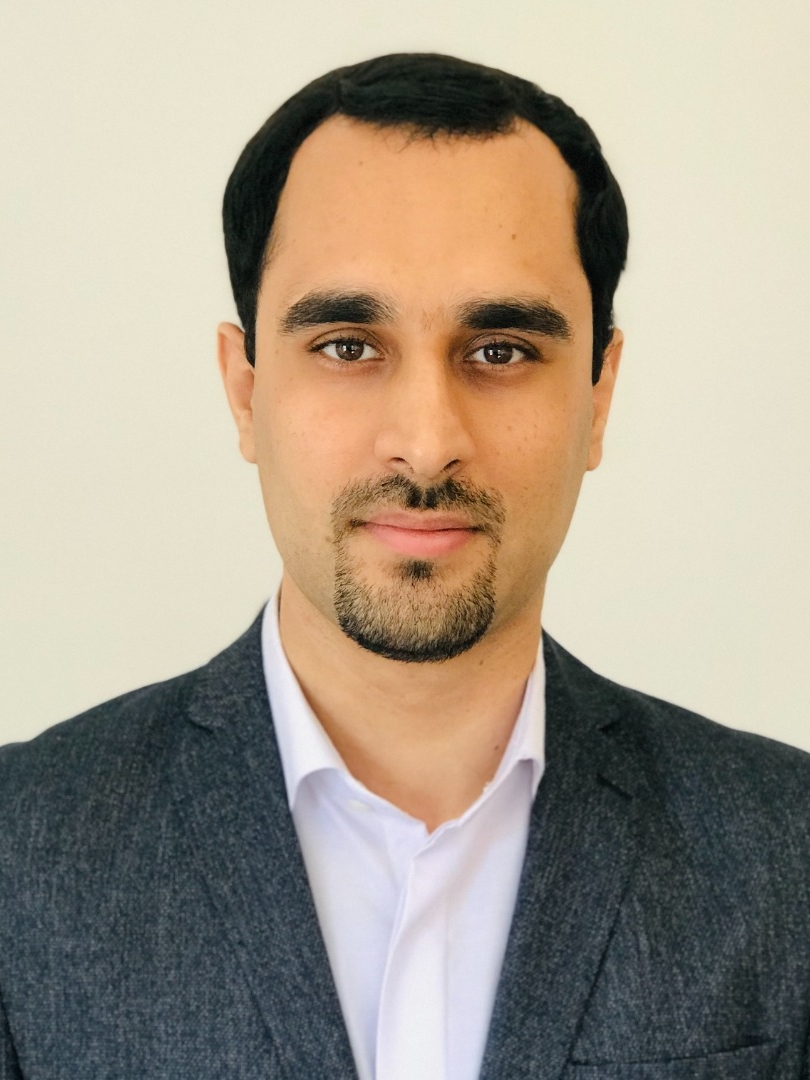}}]{Mohsen Ansari}received the Ph.D. degree in computer engineering from the Sharif University of Technology, Tehran, Iran, in 2021. He is currently an Assistant Professor of computer engineering at the Sharif University of Technology, Tehran, Iran. He was a visiting researcher with the chair for Embedded Systems (CES), Karlsruhe Institute of Technology (KIT), Germany, from 2019 to 2021. His research interests include cyber-physical systems, embedded machine learning, edge, fog, and cloud computing, and thermal and low-power design of CPSs.
\end{IEEEbiography}

\begin{IEEEbiography}[{\includegraphics[width=1in,height=1.25in, clip,keepaspectratio]{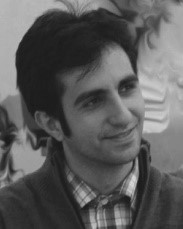}}]{MohammadAmin Fazli}
received the B.Sc. degree in hardware engineering and the M.Sc. and Ph.D. degrees in software engineering from the Sharif University of Technology, Tehran, Iran, in 2009, 2011, and 2015, respectively. He is currently an Assistant Professor at the Sharif University of Technology, where he is also a Research and Development Supervisor with the Intelligent Information Solutions Center. His current research interests include game theory, combinatorial optimization, computational business and economics, graphs and combinatorics, complex networks, and dynamical systems.
\end{IEEEbiography}

\begin{IEEEbiography}[{\includegraphics[width=1in,height=1.25in, clip,keepaspectratio]{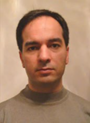}}]{Alireza Ejlali} received the Ph.D. degree in computer engineering from the Sharif University of Technology (SUT), Tehran, Iran, in 2006, where he is currently an Associate Professor of computer engineering. From 2005 to 2006, he was a Visiting Researcher with the Electronic Systems Design Group, University of Southampton, Southampton, U.K. He is currently the Director of the Embedded Systems Research Laboratory with the Department of Computer Engineering, Sharif University of Technology. His research interests include low-power design, real-time, and fault-tolerant embedded systems.
\end{IEEEbiography}

\begin{IEEEbiography}[{\includegraphics[width=1in,height=1.25in, clip,keepaspectratio]{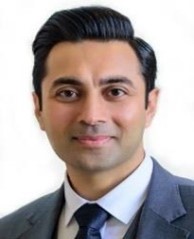}}]{Muhammad Shafique} (Senior Member, IEEE) received the Ph.D. degree in computer science from Karlsruhe Institute of Technology, Karlsruhe, Germany, in 2011. He was a Full Professor at the Institute of Computer Engineering, TU Wien, Vienna, Austria, from October 2016 to August 2020. Since September 2020, he has been with the Division of Engineering, New York University Abu Dhabi, Abu Dhabi, UAE, and is a Global Network Faculty Member of the NYU Tandon School of Engineering, Brooklyn, NY, USA. His research interests are in system-level design for brain-inspired computing, AI/machine learning hardware, wearables, autonomous systems, energy-efficient and robust computing, IoT, and smart CPS. Dr. Shafique received the 2015 ACM/SIGDA Outstanding New Faculty Award, the AI 2000 Chip Technology Most Influential Scholar Award in 2020, 2022, and 2023, six gold medals, and several best paper awards and nominations. He has given several keynotes, talks, and tutorials and organized special sessions at premier venues. He has served as the PC chair, general chair, track chair, and PC member for several conferences.
\end{IEEEbiography}

\begin{IEEEbiography}[{\includegraphics[width=1in,height=1.25in, clip,keepaspectratio]{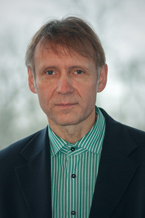}}]{Jörg Henkel} (Fellow, IEEE) received the Diploma and Ph.D. degrees (summa cum laude) from the Technical University of Braunschweig. He is currently the Chair Professor of embedded systems with the Karlsruhe Institute of Technology (KIT). Before that, he was a Research Staff Member with NEC Laboratories, Princeton, NJ, USA. His research interests include co-design for embedded hardware/software systems with respect to power security and means of embedded machine learning. He is the Vice President of Publications at IEEE CEDA and a fellow of the ACM. He has led several conferences as the General Chair, including ICCAD and ESWeek, and is currently the General of DAC’60. He serves as a steering committee chair/member for leading conferences and journals for embedded and cyber-physical systems. He has coordinated the DFG Program SPP 1500 “Dependable Embedded Systems” and is a Site Coordinator of the DFG TR89 Collaborative Research Center on “Invasive Computing.” He is the Chairman of the IEEE Computer Society, Germany Chapter. He has received six best paper awards throughout his career from, among others, ICCAD, ESWeek, and DATE. For two consecutive terms each, he served as the Editor-in-Chief for both the ACM Transactions on Embedded Computing Systems and the IEEE Design \& Test magazine.
\end{IEEEbiography}

\vfill

\end{document}